\documentclass[journal]{IEEEtran}
\usepackage{amsmath,amsfonts}
\usepackage{algorithmic}
\usepackage{algorithm}
\usepackage{array}
\usepackage[caption=false,font=normalsize,labelfont=sf,textfont=sf]{subfig}
\usepackage{textcomp}
\usepackage{stfloats}
\usepackage{url}
\usepackage{verbatim}
\usepackage{graphicx}
\usepackage{cite}
\usepackage{lineno}
\usepackage{multirow}
\usepackage[pagebackref,breaklinks,colorlinks]{hyperref}
\usepackage{booktabs}
\usepackage{xcolor}
\usepackage{tablefootnote}
\usepackage{tabularx}
\usepackage{lscape}
\usepackage{afterpage}
\usepackage{geometry}

\usepackage{amssymb}
\usepackage{geometry}
\usepackage{tabularray}
\definecolor{mygreen}{HTML}{32963d}

\newcommand*{\mline}[2]{%
\begingroup
    \renewcommand*{\arraystretch}{1.1}%
   \begin{tabular}[c]{@{}>{\raggedright\arraybackslash}p{#1}@{}}#2\end{tabular}%
  \endgroup
}

\def\dataw{8cm}
\def\trainw{9cm}

\begin{document}

\title{Second FRCSyn-onGoing:\\Winning Solutions and Post-Challenge Analysis to Improve Face Recognition with Synthetic Data}

\author{Ivan DeAndres-Tame, Ruben Tolosana, Pietro Melzi, Ruben Vera-Rodriguez, Minchul Kim, Christian Rathgeb, Xiaoming Liu, Luis F. Gomez, Aythami Morales, Julian Fierrez, Javier Ortega-Garcia, Zhizhou Zhong, Yuge Huang, Yuxi Mi, Shouhong Ding, Shuigeng Zhou, Shuai He, Lingzhi Fu, Heng Cong, Rongyu Zhang, Zhihong Xiao, Evgeny Smirnov, Anton Pimenov, Aleksei Grigorev, Denis Timoshenko, Kaleb Mesfin Asfaw, Cheng Yaw Low, Hao Liu, Chuyi Wang, Qing Zuo, Zhixiang He, Hatef Otroshi Shahreza, Anjith George, Alexander Unnervik, Parsa Rahimi, S\'{e}bastien Marcel, Pedro C. Neto, Marco Huber, Jan Niklas Kolf, Naser Damer, Fadi Boutros, Jaime S. Cardoso, Ana F. Sequeira, Andrea Atzori, Gianni Fenu, Mirko Marras, Vitomir \v{S}truc, Jiang Yu, Zhangjie Li, Jichun Li, Weisong Zhao, Zhen Lei, Xiangyu Zhu, Xiao-Yu Zhang, Bernardo Biesseck, Pedro Vidal, Luiz Coelho, Roger Granada, David Menotti   

\thanks{
2nd FRCSyn Challenge Organizers: Ruben Tolosana, Ivan DeAndres-Tame, Pietro Melzi, Ruben Vera-Rodriguez, Minchul Kim. Christian Rathgeb, Xiaoming Liu, Aythami Morales, Julian Fierrez and Javier Ortega-Garcia. Information related to each author is included at the end of the article. Link to 2nd FRCSyn Challenge: \href{https://frcsyn.github.io/CVPR2024.html}{https://frcsyn.github.io/CVPR2024.html}}}




\maketitle

\begin{abstract}
Synthetic data is gaining increasing popularity for face recognition technologies, mainly due to the privacy concerns and challenges associated with obtaining real data, including diverse scenarios, quality, and demographic groups, among others. It also offers some advantages over real data, such as the large amount of data that can be generated or the ability to customize it to adapt to specific problem-solving needs. To effectively use such data, face recognition models should also be specifically designed to exploit synthetic data to its fullest potential. In order to promote the proposal of novel Generative AI methods and synthetic data, and investigate the application of synthetic data to better train face recognition systems, we introduce the 2$^\text{nd}$ FRCSyn-onGoing challenge, based on the 2$^\text{nd}$ Face Recognition Challenge in the Era of Synthetic Data (FRCSyn), originally launched at CVPR 2024. This is an ongoing challenge that provides researchers with an accessible platform to benchmark \textit{i)} the proposal of novel Generative AI methods and synthetic data, and \textit{ii)} novel face recognition systems that are specifically proposed to take advantage of synthetic data. We focus on exploring the use of synthetic data both individually and in combination with real data to solve current challenges in face recognition such as demographic bias, domain adaptation, and performance constraints in demanding situations, such as age disparities between training and testing, changes in the pose, or occlusions. Very interesting findings are obtained in this second edition, including a direct comparison with the first one, in which synthetic databases were restricted to DCFace and GANDiffFace.
\end{abstract}


\section{Introduction}\label{sec:intro}

Face biometrics is a very popular area within Computer Vision and Pattern Recognition, finding applications across various domains such as person recognition \cite{wang2021deep, du2022elements}, healthcare \cite{bisogni2022impact, gomez2021improving}, and e-learning \cite{daza2023matt}, among others. In recent years, with the fast development of deep learning, significant advances have been made in areas like face recognition (FR) \cite{deng2019arcface, kim2022adaface}, surpassing previous benchmarks. However, FR technology still faces challenges in several research directions, including explainability \cite{deandrestame2024how, crum2023explain, shen2022interfacegan}, demographic bias \cite{terhoerst2021comprehensive, melzi2023synthetic}, privacy \cite{morales2021sensitivenets, melzi2023multi, melzi2024overview}, and robustness against adverse conditions \cite{kim2022adaface}, such as aging~\cite{zhao2022towards}, pose variations~\cite{valle2021multi, tran2019representation}, illumination changes~\cite{mudunuri2016low}, and occlusions~\cite{qiu2022end2end}.

\begin{figure*}[t]
    \centering
    \includegraphics[width=\textwidth]{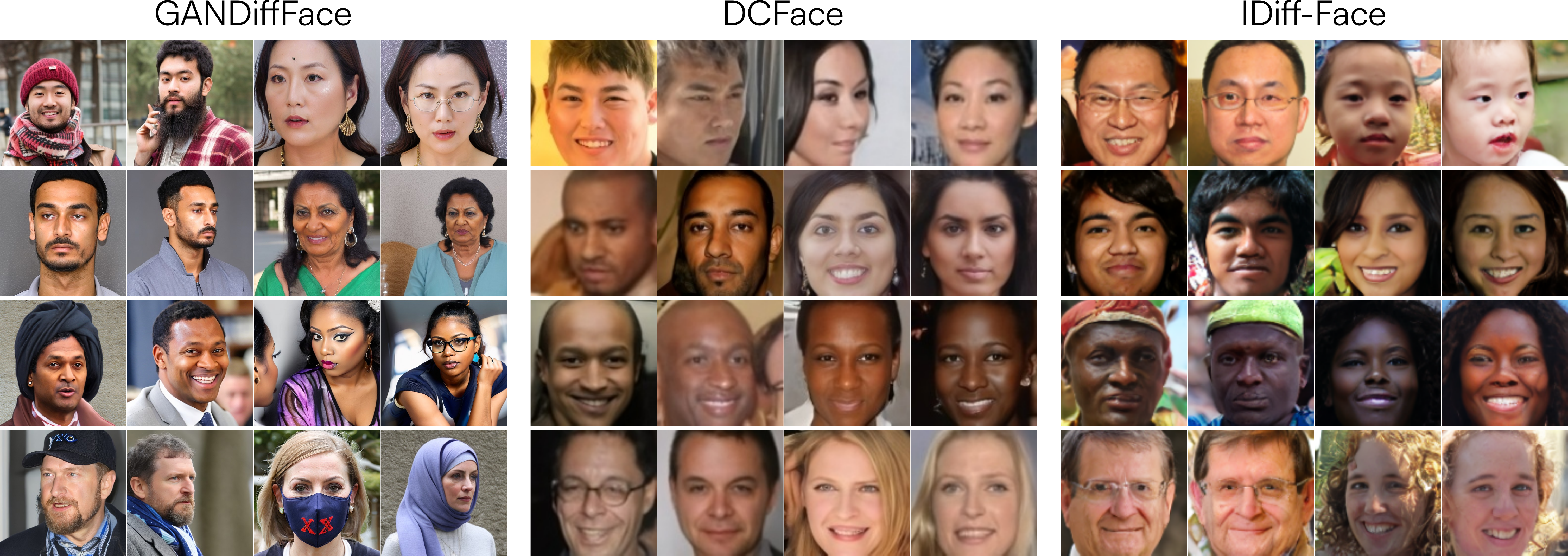}
    \caption{Examples of synthetic identities and variations for different demographic groups using GANDiffFace~\cite{melzi2023gandiffface} (left), DCFace~\cite{kim2023dcface} (middle) and IDiff-Face~\cite{boutros2023idiff} (right).}
    \label{fig:gandiff_ex}
\end{figure*}

Synthetic data has gained popularity as a good solution to mitigate some of these drawbacks~\cite{boutros2023synthetic, joshi2024synthetic}, allowing the generation of \textit{i)} a large number of facial images from different non-existent identities, and \textit{ii)} variability in terms of demographic attributes and scenario conditions. In this context, we refer to demographics as whole societies or smaller groups defined by criteria such as ethnicity, sex, or age~\cite{terhoerst2021comprehensive, sun2018demographic}. Several Generative AI approaches have been presented in the last couple of years for the synthesis of face images, considering state-of-the-art deep learning methods. The most popular approaches to generate synthetic facial images are Generative Adversarial Networks (GANs)~\cite{goodfellow2014generative}, Diffusion Models~\cite{ho2020denoising} or the combination of both~\cite{melzi2023gandiffface}. However, there are other less common approaches that do not rely on these models and still achieve a high level of realism in image generation~\cite{bae2023digiface, zhang2023iti}. Some examples of synthetic face images generated using some of these methods are shown in Figure~\ref{fig:gandiff_ex}.

Beyond the generation of synthetic faces, another critical aspect lies in understanding the potential applications and benefits of synthetic data in enhancing FR technology. Recent studies have highlighted a performance gap between FR systems trained only with synthetic data and those trained on real data \cite{kim2023dcface, qiu2021synface}. Nevertheless, the results achieved in the 1$^\text{st}$ edition of the Face Recognition Challenge in the Era of Synthetic Data (FRCSyn)~\cite{melzi2024frcsyn, melzi2024frcsyna}, emphasize the relevance of synthetic data, either alone or merged with real data, in mitigating challenges in FR, such as demographic bias \cite{melzi2024frcsyn, melzi2024frcsyna}. Notably, in the 1$^\text{st}$ FRCSyn-onGoing, only synthetic data from DCFace \cite{kim2023dcface} and GANDiffFace \cite{melzi2023gandiffface} methods were allowed for training FR systems. Additionally, together with novel generative methods, improving FR technology involves refining the design and training processes to address domain gaps between real and synthetic data in certain scenarios. For instance, observations from the 1$^\text{st}$ FRCSyn-onGoing revealed that most teams considered similar deep learning architectures (e.g., ResNet-100 \cite{he2016deep}) and loss functions (e.g., AdaFace \cite{kim2022adaface}), commonly used in FR systems trained with real data. Moreover, the use of synthetic facial data is not limited only to FR. With the emerging popularity of foundational models~\cite{radford2021learning, oquab2023dinov2}, synthetic facial data can also be leveraged to provide these large models with a general understanding of what a human face looks like, serving as pretraining for many other tasks~\cite{wang2023hulk, tang2023humanbench, khirodkar2025sapiens}. Additionally, the fact that these data can be labeled as they are generated allows the creation of novel datasets for various tasks such as attribute detection, facial expression recognition, and more. Although in the field of synthetic faces there are still not many public datasets in these lines of research, several studies have already highlighted the benefits of synthetic data for biometric tasks such as facial expression recognition~\cite{bozorgtabar2019using}, signature verification~\cite{tolosana2021deepwritesyn}, action recognition~\cite{hwang2023eldersim}, or pose estimation~\cite{varol2017learning}.

In order to promote the development of novel face generative methods and the creation of synthetic face databases, as well as investigate the application of synthetic data to better train FR systems, we have organized the 2$^\text{nd}$ FRCSyn-onGoing Challenge, which is based on the 2$^\text{nd}$ FRCSyn Challenge as part of CVPR 2024\footnote{\href{https://frcsyn.github.io/CVPR2024.html}{https://frcsyn.github.io/CVPR2024.html}}~\cite{deandres2024second}. In this 2$^\text{nd}$ edition, we introduce new sub-tasks allowing participants to train FR systems using synthetic data generated with their preferred generative frameworks, offering more flexibility compared to the 1$^\text{st}$ edition \cite{melzi2024frcsyna, melzi2024frcsyn}. Additionally, new sub-tasks with varied experimental settings are included to explore how FR systems can be trained under both constrained and unconstrained scenarios regarding the amount of synthetic training data. The FRCSyn Challenge aims to address the following research questions:
\begin{enumerate}
\item What are the limitations of FR technology trained only with synthetic data?
\item Can synthetic data help alleviate current limitations in FR technology?
\end{enumerate}
These questions have become increasingly relevant after the discontinuation of popular real FR databases due to privacy concerns\footnote{\href{https://exposing.ai/about/news/}{https://exposing.ai/about/news/} (March, 2024)} and the introduction of new regulatory laws\footnote{\href{https://artificialintelligenceact.eu}{https://artificialintelligenceact.eu} (March, 2024)}.

The foundation of the present article was established in an earlier publication~\cite{deandres2024second}, with the current version notably extending it through: \textit{i)} a more extensive description and analysis of the top synthetic face generation methods and FR systems presented so far in this 2$^\text{nd}$ FRCSyn-onGoing, including key graphical representations of the proposed systems to improve the understanding of the reader, \textit{ii)} incorporating additional metrics in the evaluation of the proposed FR systems in order to analyze different operational scenarios, \textit{iii)} presenting an in-depth analysis of the performance achieved for various demographic groups and databases used for evaluation, together with novel figures and tables, and \textit{iv)} a direct comparison between the results obtained in this 2$^\text{nd}$ edition and the ones obtained in 1$^\text{st}$ edition~\cite{melzi2024frcsyn, melzi2024frcsyna}, highlighting very interesting findings.

The remainder of the article is organized as follows. Section~\ref{sec:ddbb} describes the databases considered at the 2$^\text{nd}$ FRCSyn-onGoing. Section~\ref{sec:setup} explains the experimental setup of the challenge, including the different tasks and sub-tasks, the experimental protocol, metrics,  and restrictions. In Section~\ref{sec:sys}, we describe the approaches proposed by the top-6 participating teams. Section~\ref{sec:results} presents the best results achieved so far in the different tasks and sub-tasks of 2$^\text{nd}$ FRCSyn-onGoing, emphasizing the key results of the challenge. Finally, in Section \ref{sec:conclusion}, we provide some conclusions, highlighting potential future research directions in the field.

\begin{table*}[t]
\centering
\caption{Description of some possible generative methods and synthetic databases that can be used by participants in the 2$^\text{nd}$ FRCSyn-onGoing. Id = Identities, Img = Images}
\label{tab:ddbb}
\resizebox{0.7\linewidth}{!}{\begin{tabular}{llccc}
\hline
\textbf{Database} & \textbf{Framework} & \textbf{\# Id} & \textbf{\# Img/Id} & \textbf{\# Img} \\ \hline
\multirow{2}{*}{DCFace~\cite{kim2023dcface}} & \multirow{2}{*}{Diffusion Model} & 20K & 50 & \multirow{2}{*}{1,200K} \\
 &  & 40K & 5 &  \\ \hline
\multirow{2}{*}{GANDiffFace~\cite{melzi2023gandiffface}} & GAN & \multirow{2}{*}{10K} & \multirow{2}{*}{50} & \multirow{2}{*}{500K} \\
 & Diffusion Model &  &  &  \\ \hline
\begin{tabular}[c]{@{}l@{}} IDiff-face\\Uniform\end{tabular}~\cite{boutros2023idiff} & Diffusion Model & 10K & 50 & 500K \\\hline
\begin{tabular}[c]{@{}l@{}} IDiff-face\\Two-Stage\end{tabular}~\cite{boutros2023idiff} & Diffusion Model & 10K & 50 & 500K \\ \hline
\multirow{2}{*}{DigiFace-1M~\cite{bae2023digiface}} & \multirow{2}{*}{3D-model} & 10K & 72 & 1.2M \\
 &  & 100K & 5 &  \\ \hline
ID3PM~\cite{kansy2023controllable} & Diffusion Model & - & - & - \\ \hline
SFace~\cite{boutros2022sface} & GAN & 10K & 60 & 0.6M \\ \hline
SYNFace~\cite{qiu2021synface} & GAN & 10K & 100 & 1M \\ \hline
ITI-GEN~\cite{zhang2023iti} & CLIP & - & - & - \\ \hline
\end{tabular}}
\end{table*}
\section{Second FRCSyn-onGoing: Databases}\label{sec:ddbb}

\subsection{Synthetic Databases} \label{sec:syn}
One of the main novelties of the 2$^\text{nd}$ FRCSyn-onGoing is that there are no restrictions in terms of the generative methods used to create synthetic data. Unlike the 1$^\text{st}$ FRCSyn-onGoing, where only synthetic data created using DCFace~\cite{kim2023dcface} and GANDiffFace~\cite{melzi2023gandiffface} was available, in this 2$^\text{nd}$ edition we allow participants to use any generative framework of their choice to create synthetic data, limiting in some sub-tasks the number of synthetic face images used to train the FR systems (more details in Section~\ref{sec:task}). As a reference, after the registration in the challenge, we provide all the participants with a list of possible state-of-the-art generative frameworks. For completeness, we summarize next and in Table~\ref{tab:ddbb} the most popular approaches available at the beginning of the challenge:

\begin{itemize}
    \item \textbf{DCFace}\footnote{\href{https://github.com/mk-minchul/dcface}{https://github.com/mk-minchul/dcface}}~\cite{kim2023dcface}: This framework is entirely based on Diffusion models, composed of a sampling stage for the generation of synthetic identities $X_{ID}$, and a mixing stage for the generation of images $X_{ID, sty}$ with the same identities $X_{ID}$ from the sampling stage and the style selected from a ``style bank'' of images $X_{sty}$.
    
    \item \textbf{GANDiffFace}\footnote{\href{https://github.com/PietroMelzi/GANDiffFace}{https://github.com/PietroMelzi/GANDiffFace}}~\cite{melzi2023gandiffface}: This framework combines StyleGAN~\cite{karras2021alias} and a Diffusion Model, \textit{i.e.,} DreamBooth~\cite{ruiz2023dreambooth}, to generate fully synthetic FR databases with desired properties such as human face realism, controllable demographic distributions, and realistic intra-class variations (\textit{e.g.,} changes in pose, expression, and occlusions). Graphical examples are shown in Figure~\ref{fig:gandiff_ex}.

    \item \textbf{IDiff-Face}\footnote{\href{https://github.com/fdbtrs/IDiff-Face}{https://github.com/fdbtrs/IDiff-Face}}~\cite{boutros2023idiff}: This framework uses a Diffusion Model conditioned on identity context, which allows the model to either generate variations of existing authentic images by using authentic embeddings or to generate novel synthetic identities by using synthetic face embeddings. The authors presented two distinct datasets: one by generating identity context in a two-stage process, and the other through a synthetic uniform representation.
    
    \item \textbf{DigiFace-1M}\footnote{\href{https://github.com/microsoft/DigiFace1M}{https://github.com/microsoft/DigiFace1M}}~\cite{bae2023digiface}: This framework can generate large-scale synthetic face images with many unique subjects based on 3D parametric model rendering. It considers the method introduced by Wood \textit{et al.}~\cite{wood2021fake}, tackling the ethical and labeling problems associated with the generation of synthetic data. 
    
    \item \textbf{ID3PM}~\cite{kansy2023controllable}: This framework considers a Diffusion Model to perform an inversion of a FR model generating new images from Gaussian noise with various backgrounds, lighting, poses, and expressions while preserving the identity.
    
    \item \textbf{SFace}\footnote{\href{https://github.com/fdbtrs/SFace-Privacy-friendly-and-Accurate-Face-Recognition-using-Synthetic-Data}{https://github.com/fdbtrs/SFace-Privacy-friendly-and-Accurate-Face-Recognition-using-Synthetic-Data}}~\cite{boutros2022sface}: This framework uses a conditional GAN to synthetically generate face images with an adaptive discriminator augmentation to increase the diversity of the training database.
    
    \item \textbf{SYNFace}\footnote{\href{https://github.com/haibo-qiu/SynFace}{https://github.com/haibo-qiu/SynFace}}~\cite{qiu2021synface}: This framework uses DiscoFaceGAN~\cite{deng2020disentangled} to generate face images with different identities from a Mixup Face Generator.
    
    \item \textbf{ITI-GEN}\footnote{\href{https://github.com/humansensinglab/ITI-GEN}{https://github.com/humansensinglab/ITI-GEN}}\cite{zhang2023iti}: This framework uses CLIP~\cite{radford2021learning} to generate embeddings to translate the visual attribute differences into natural language differences and perform a Text-to-Image generation that is inclusive.
\end{itemize}
\begin{table*}[t]
\centering
\scriptsize
\caption{Description of the real databases considered for evaluation in the 2$^\text{nd}$ FRCSyn-onGoing.\\Id = Identities, Img = Images}
\label{tab:ddbb_real}
\resizebox{0.65\linewidth}{!}{\begin{tabular}{lllll}
\hline
\textbf{Database} & \textbf{Framework} & \textbf{\# Id} & \textbf{\# Img/Id} & \textbf{\# Img} \\ \hline
CASIA-WebFace~\cite{yi2014learning} & Real & 10.5K & 47 & 500K \\
BUPT-BalancedFace~\cite{wang2020mitigating} & Real & 24K & 45 & 1M \\
AgeDB~\cite{moschoglou2017agedb} & Real & 570 & 29 & 17K \\
CFP-FP~\cite{sengupta2016frontal} & Real & 500 & 14 & 7K \\
ROF~\cite{erakιn2021recognizing} & Real & 180 & 31 & 6K \\ \hline
\end{tabular}}
\end{table*}

\begin{table*}[h]
    \centering
    \caption{Tasks and sub-tasks for the 2$^\text{nd}$ FRCSyn-onGoing and their respective metrics and databases.\\TO = Trade-Off. GAP = Gap to Real. AVG = Average Verification Accuracy. SD = Standard Deviation. FLOPs =  Floating Point Operations Per Second. SYN = Accuracy Proposed Model. REAL = Accuracy Baseline Model. ID = Identity.}
    \label{tab:2}
    \begin{tabular}{ll} 
    \hline
    \multicolumn{2}{l}{\textbf{Task 1:} synthetic data for \textbf{demographic bias mitigation}} \\
    \multicolumn{2}{l}{\quad \textbf{Baseline}: training with only CASIA-WebFace \cite{yi2014learning}.} \\
    \multicolumn{2}{l}{\quad \textbf{Ranking}: Trade-Off, see Section \ref{sec:metrics} for more details.} \\
    \multirow{2}{*}{\quad\textbf{Metrics:}} & $ TO = AVG - SD $ \\
    & $ GAP = (REAL-SYN)/SYN $ \\\hline
    \multicolumn{2}{l}{\textbf{\textcolor{blue}{Sub-Task 1.1: [\textcolor{mygreen}{constrained}]}} training exclusively with \textbf{synthetic} data } \\
    \multicolumn{2}{l}{\quad \textbf{Train}: maximum $500K$ face images (\textit{e.g.,} $10K$ IDs and $50$ images per ID). } \\
    \multicolumn{2}{l}{\quad \textbf{Eval}: BUPT-BalancedFace \cite{wang2020mitigating}. } \\ \hline

    \multicolumn{2}{l}{\textbf{\textcolor{blue}{Sub-Task 1.2: [\textcolor{mygreen}{unconstrained}]}} training exclusively with \textbf{synthetic} data } \\
    \multicolumn{2}{l}{\quad \textbf{Train}: no restrictions in terms of the number of face images. } \\
    \multicolumn{2}{l}{\quad \textbf{Eval}: BUPT-BalancedFace. } \\ \hline
 
    \multicolumn{2}{l}{\textbf{\textcolor{blue}{Sub-Task 1.3: [\textcolor{mygreen}{constrained}]}} training with \textbf{real and synthetic} data } \\ 
    \multicolumn{2}{l}{\quad \textbf{Train}: CASIA-WebFace, and maximum $500K$ face synthetic images. } \\ 
    \multicolumn{2}{l}{\quad \textbf{Eval}: BUPT-BalancedFace. } \\ \hline \hline

    \multicolumn{2}{l}{\textbf{Task 2:} synthetic data for \textbf{overall performance improvement} } \\
    \multicolumn{2}{l}{\quad \textbf{Baseline}: training with only CASIA-WebFace. } \\
    \multicolumn{2}{l}{\quad \textbf{Ranking}: average accuracy, see Section \ref{sec:metrics} for more details.} \\ 
    \multirow{2}{*}{\quad \textbf{Metrics:}} & $ AVG $ \\
    & $ GAP = (REAL-SYN)/SYN $ \\\hline

    \multicolumn{2}{l}{\textbf{\textcolor{blue}{Sub-Task 2.1: [\textcolor{mygreen}{constrained}]}} training with only \textbf{synthetic} data } \\ 
    \multicolumn{2}{l}{\quad \textbf{Train}: maximum $500K$ face images.} \\
    \multicolumn{2}{l}{\quad \textbf{Eval}: BUPT-BalancedFace, AgeDB~\cite{moschoglou2017agedb}, CFP-FP~\cite{sengupta2016frontal}, and ROF~\cite{erakιn2021recognizing}. } \\ \hline

    \multicolumn{2}{l}{\textbf{\textcolor{blue}{Sub-Task 2.2: [\textcolor{mygreen}{unconstrained}]}} training with only \textbf{synthetic} data } \\ 
    \multicolumn{2}{l}{\quad \textbf{Train}: no restrictions in terms of the number of face images. } \\
    \multicolumn{2}{l}{\quad \textbf{Eval}: BUPT-BalancedFace, AgeDB, CFP-FP, and ROF. } \\  \hline

    \multicolumn{2}{l}{\textbf{\textcolor{blue}{Sub-Task 2.3: [\textcolor{mygreen}{constrained}]}} training with \textbf{real and synthetic} data } \\
    \multicolumn{2}{l}{\quad \textbf{Train}: CASIA-WebFace, and maximum $500K$ face synthetic images. } \\
    \multicolumn{2}{l}{\quad \textbf{Eval}: BUPT-BalancedFace, AgeDB, CFP-FP, and ROF. } \\ \hline \hline
    \multirow{3}{*}{\textbf{Restrictions:}} & FLOPS$\leq$50 GFLOPS \\
    & Only the specified databases can be used for training. \\
    & Generative models cannot be used to generate supplementary data. \\ \hline
    \end{tabular}
\end{table*}
These are just some possible generative frameworks, with the corresponding synthetic databases available, that can be used by participants. But, as indicated before, the purpose of the 2$^\text{nd}$ FRCSyn-onGoing is to promote the proposal of novel generative methods and the creation of better synthetic databases to improve the performance of FR systems. It is important to mention that in the 2$^\text{nd}$ FRCSyn-onGoing, synthetic data is exclusively used in the training stage of FR technology, replicating realistic operational scenarios.

\subsection{Real Databases}\label{sec:real_bbdd}

For the training of the FR systems participants are allowed to use only the \textbf{CASIA-WebFace}~\cite{yi2014learning} as real data (depending on the sub-task, please see Section~\ref{sec:task}). This database contains $494,414$ face images of $10,575$ real identities collected from the web. For the final evaluation of the proposed FR systems, we consider the same four real databases used at the 1$^\text{st}$ FRCSyn Challenge~\cite{melzi2024frcsyn, melzi2024frcsyna}, as they consider key challenges in FR such as demographic bias, pose variations, aging, and occlusions. We summarize next and in Table~\ref{tab:ddbb_real} each of them:

\begin{itemize}
    \item \textbf{BUPT-BalancedFace}~\cite{wang2020mitigating} is designed to address performance disparities across different ethnic groups. We relabel it according to the FairFace classifier \cite{karkkainen2021fairface}, which provides labels for ethnicity (White, Black, Asian, Indian) and gender (Male, Female). We then consider the eight demographic groups obtained from all possible combinations of four ethnic groups and genders. We are aware that these groups do not comprehensively represent the entire spectrum of real world ethnic diversity. Nevertheless, the selection of these categories, while imperfect, is primarily driven by the need to align with the demographic categorizations used in BUPT-BalancedFace to facilitate easier and more consistent evaluation. 

    \item \textbf{AgeDB}~\cite{moschoglou2017agedb} contains facial images featuring the same subjects at different ages in different environmental contexts.
    
    \item \textbf{CFP-FP}~\cite{sengupta2016frontal} presents facial images from subjects with great changes in pose, with both frontal and profile images, and different environmental contexts. 
    
    \item \textbf{ROF}~\cite{erakιn2021recognizing} consists of occluded faces with both upper face occlusion, due to sunglasses, and lower face occlusion, due to masks.
\end{itemize}

Finally, it is important to highlight that, as different databases are considered for training and evaluation, we also intend to analyze the generalization ability of the proposed FR systems.

\section{Second FRCSyn-onGoing: Setup}\label{sec:setup}

Due to the success of the 1$^\text{st}$ FRCSyn-onGoing~\cite{melzi2024frcsyn, melzi2024frcsyna}, we also decided to run the 2$^\text{nd}$ edition in Codalab\footnote{\href{https://codalab.lisn.upsaclay.fr/competitions/16970}{https://codalab.lisn.upsaclay.fr/competitions/16970}}, an open-source framework designed for conducting scientific competitions and benchmarks. On this platform, participants can find the competition's requirements and limitations and can submit their scores to automatically obtain \textit{i)} the evaluation metrics of their system, and \textit{ii)} the position on the challenge leaderboard. Table~\ref{tab:2} provides an overview of the key aspects of the experimental protocol, metrics and restrictions for each sub-task. More detailed explanations can be found in their respective subsections.

\subsection{Tasks}\label{sec:task}

Similar to the 1$^\text{st}$ FRCSyn-onGoing~\cite{melzi2024frcsyna,melzi2024frcsyn}, in this 2$^\text{nd}$ edition we also explore the application of synthetic data for training FR systems, with a specific focus on addressing two critical aspects in current FR technology: \textit{i)} mitigating demographic bias, and \textit{ii)} enhancing overall performance under challenging conditions that include variations in age and pose, the presence of occlusions, and diverse demographic groups. To investigate these two areas, we consider two different tasks, each comprising three sub-tasks. Each sub-task considers different types (real/synthetic) and amounts of data for training the FR systems. Consequently, the 2$^\text{nd}$ edition comprises 6 different sub-tasks.

\paragraph*{\textbf{Task 1}}

The first task focuses on using synthetic data to mitigate demographic biases within FR systems. To evaluate the performance of these systems, we create sets of mated and non-mated comparisons using subjects from the BUPT-BalancedFace database~\cite{wang2020mitigating}. We consider the eight demographic groups defined in Section~\ref{sec:real_bbdd}, which result from the combination of four ethnicities (White, Black, Asian, and Indian) and two genders (Male and Female), ensuring a balanced representation across these groups in the comparison lists. For non-mated comparisons, we exclusively pair subjects within the same demographic group, as these hold greater relevance compared to non-mated comparisons involving subjects from different demographic groups.

\paragraph*{\textbf{Task 2}}
The second proposed task focuses on using synthetic data to enhance the overall performance of FR systems under challenging conditions. To assess the effectiveness of the proposed systems, we use lists of mated and non-mated comparisons selected from subjects from the different evaluation databases, each one designed to address specific challenges in FR. Specifically, BUPT-BalancedFace is used to consider diverse demographic groups, whereas AgeDB, CFP-FP, and ROF to assess age, pose, and occlusion challenges respectively.

\subsection{Experimental protocol}
\paragraph*{\textbf{Training}}
The six sub-tasks introduced in the 2$^\text{nd}$ FRCSyn-onGoing are mutually independent. This implies that participants have the flexibility to participate in any number of sub-tasks based on their preferences. For each selected sub-task, participants are required to develop a FR system and train it twice: \textit{i)} using the authorized real databases exclusively, \textit{i.e.,} CASIA-WebFace \cite{yi2014learning}, and \textit{ii)} following the specific requirements of the chosen sub-task, as summarized in Table \ref{tab:2}. According to this protocol, participants must provide both the \textit{baseline system} and the \textit{proposed system} for each specific sub-task. The baseline system plays a critical role in evaluating the impact of synthetic data on training and serves as a reference point for comparing the proposed model against the conventional practice of training only with real databases. To maintain consistency, the baseline FR system, trained exclusively with real data, and the proposed FR system, trained according to the specifications of the selected sub-task, must have the same architecture and training protocol.

\paragraph*{\textbf{Evaluation}}
In each sub-task, participants received the comparison files comprising both mated and non-mated comparisons, which are used to evaluate the performance of their proposed FR systems. Task 1 involves a single comparison file containing balanced comparisons of different demographic groups of the BUPT~\cite{wang2020mitigating} database, while Task 2 comprises four comparison files, each corresponding to every specific real-world databases considered (\textit{i.e.,} BUPT, AgeDB~\cite{moschoglou2017agedb}, CFP-FP~\cite{sengupta2016frontal}, and ROF~\cite{erakιn2021recognizing}). During the evaluation of each sub-task, participants are required to submit two files per database through the Codalab platform: \textit {i)} the scores of the baseline system, and \textit{ii)} the scores of the proposed system. Finally, for each sub-task, participants must submit a file including the decision threshold for each FR system (\textit{i.e.,} baseline and proposed). The submitted scores must fall within the range of $[0, 1]$, with lower scores indicating non-mated comparisons, and vice versa.

\subsection{Evaluation Metrics}\label{sec:metrics}
We evaluate the FR systems using a protocol based on lists of mated and non-mated comparisons for each sub-task and database. From the scores and thresholds provided by participants, we calculate the binary decision and the verification accuracy. Additionally, we calculate the gap to real (GAP) \cite{kim2023dcface} as follows: $\text{GAP} = \left( \text{REAL} - \text{SYN} \right)/\text{SYN}$, with $\text{REAL}$ representing the verification accuracy of the baseline system and $\text{SYN}$ the verification accuracy of the proposed system, trained with synthetic (or real + synthetic) data. Other metrics such as False Non-Match Rate (FNMR) at a fixed operational point, or the Area Under the ROC Curve which are very popular for the analysis of FR systems in real-world applications, are also computed from the scores provided by participants. Next, we explain how participants are ranked in the different tasks.

\paragraph*{\textbf{Task 1}}
To rank participants and determine the winners of Sub-Tasks 1.1, 1.2, and 1.3, we closely examine the trade-off between the average (AVG) and standard deviation (SD) of the verification accuracy across the eight demographic groups defined in Section~\ref{sec:real_bbdd}. We define the trade-off metric (TO) as follows: $\text{TO} = \text{AVG} - \text{SD}$. This metric involves plotting the average accuracy on the x-axis and the standard deviation on the y-axis in a 2D space. Multiple 45-degree parallel lines are drawn to identify the winning team, whose performance is located on the far right of these lines. With this proposed metric, we reward FR systems that achieve good levels of performance and fairness simultaneously, unlike common benchmarks based only on recognition performance. The standard deviation of verification accuracy across demographic groups is a common metric for assessing bias and should be reported by any work addressing demographic bias mitigation.

\paragraph*{\textbf{Task 2}}
To rank participants and establish the winners in Sub-Tasks 2.1, 2.2, and 2.3, we examine the average verification accuracy from the four different databases designated for evaluation, as described in Section \ref{sec:ddbb}. This approach enables us to assess four main challenges of FR technologies: \textit{i)} the representation of diverse demographic groups, \textit{ii)} the impact of aging on recognition, \textit{iii)} the variations in facial pose, and \textit{iv)} the challenges made by occlusions. This evaluation provides a comprehensive overview of FR systems in real operational scenarios.

\subsection{Restrictions}
Participants have the freedom to choose the FR system for each task as long as the number of Floating Point Operations Per Second (FLOPs) of the system does not exceed 50 GFLOPs. This threshold has been established to facilitate the exploration of innovative architectures and encourage the use of diverse models while preventing the dominance of excessively large models. Participants are also free to use their preferred training modality, with the requirement that only the specified databases are used for training. Generative models cannot be used to generate supplementary data. Participants are allowed to use non-face databases for pre-training purposes and use traditional data augmentation techniques using the authorized training databases. To maintain the integrity of the evaluation process, the organizers reserve the right to disqualify participants if anomalous results are detected or if participants fail to adhere to the challenge's rules.

\section{Second FRCSyn-onGoing: Systems Description}\label{sec:sys}

In this 2$^\text{nd}$ FRCSyn-onGoing, we encourage participants to propose novel Generative AI methods for the creation of synthetic data. Besides, we also give participants the freedom to choose the FR architecture and training methods. Table~\ref{tab:data_meth} summarizes for each team the key information in terms of the proposed synthetic data and FR system. This table serves as a quick reference, while more detailed explanations of each team's approach and methodology can be found in the corresponding subsections. Teams are arranged by their average ranking in the 6 sub-tasks from the 2$^\text{nd}$ FRCSyn-onGoing. In general, we can see that most teams have decided to use synthetic data from DCFace~\cite{kim2023dcface} and IDiff-Face~\cite{boutros2023idiff} databases, improving also the original data through cleaning and selection approaches, among other more sophisticated techniques. Also, regarding the FR technologies, most of them are based on ResNet~\cite{he2016deep} and IResNet~\cite{duta2021improved} architectures, with AdaFace~\cite{kim2022adaface} and ArcFace~\cite{deng2019arcface} as the main used losses. However, some teams proposed their own methods to generate synthetic facial images, as well as to train their FR models. Next, we describe the specific details of the top-11 proposed systems in the 2$^\text{nd}$ FRCSyn-onGoing.

\newgeometry{left=0.5cm,right=0.5cm,bottom=1.5cm, top=1cm}
\begin{landscape}
\begin{table}[t]
\caption{Description of the key information in terms of the proposed synthetic data and FR system. LDM = Latent Diffusion Model, DDPM = Diffusion Probabilistic Model, DDIM = Denoising Diffusion Implicit Models, HDT = Hourglass Diffusion Transformer}
\label{tab:data_meth}
\centering
\resizebox{1.2\textwidth}{!}{
\centering
\begin{tabular}{lllclll}
\toprule
\textbf{Team}& \textbf{Country} & \begin{tabular}[c]{@{}l@{}} \textbf{Sub}\\\textbf{Task}\end{tabular} & \textbf{Fig.} & \begin{tabular}[c]{@{}l@{}}\textbf{Synthetic} \\\textbf{Database}\end{tabular} & \textbf{Synthetic Data - Team Improvements} & \textbf{FR Model - Training method} \\ \hline

ADMIS & China & All & \ref{fig:Admis} & Novel & \mline{\dataw}{They used an LDM~\cite{rombach2022high} to create the faces. The LDM uses ID embeddings as context. These contexts were generated with a DDPM~\cite{ho2020denoising}. They used cosine similarity to ensure the quality and a DDIM~\cite{song2021denoising} to accelerate the sampling. They finally oversampled the ID to enhance consistency.} & \mline{\trainw}{\strut They appliyed a 25\% dropout on the dimensions from the feature embeddings.\\\textbf{Loss:} ArcFace~\cite{deng2019arcface}\\\textbf{Backbone:} IResNet-101~\cite{duta2021improved}\strut}\\ \hline

OPDAI & China & All & \ref{fig:OPDAI} & DCFace~\cite{kim2023dcface} & \mline{\dataw}{They generated 10 new faces/ID with Photomaker~\cite{li2024photomaker} and replaced them randomly in DCFace.} & \mline{\trainw}{\strut They used different heads for different databases and calculated the final loss as the average.\\\textbf{Loss:} AdaFace~\cite{kim2022adaface}\\\textbf{Backbone:} IResNet-100\cite{duta2021improved}\strut} \\\hline

ID R\&D & USA & All & \ref{fig:IDRnD} & Novel & \mline{\dataw}{They used an HDT~\cite{crowson2024scalable} model to generate the synthetic images. It is trained with identity embeddings and style embeddings processed through a VQVAE~\cite{oord2017neural}. They also used a StyleNAT~\cite{walton2022stylenat} to generate variability from the images generated with the HDT.} & \mline{\trainw}{\strut They trained 2 models, one with color, geometric augmentations, and FaceMix-B~\cite{garaev2023facemixa} and the second with horizontal flipping augmentation. The final score was obtained by combining the outputs. \\\textbf{Loss:} UniFace~\cite{zhou2023uniface} \\ \textbf{Backbone:} IResNet-200\cite{duta2021improved}\strut } \\\hline

K-IBS-DS & S. Korea & All & \ref{fig:KIBS} & DCFace~\cite{kim2023dcface} & \multicolumn{1}{c}{-} & \mline{\trainw}{\strut They modified AdaFace following SlackedFace~\cite{low2023slackedface} by changing the initialization and replacing the L2-norm with the p-norm. They used 3 different backbones and the final score is the combination of them.\\ \textbf{Loss:} SlackedFace\cite{low2023slackedface}\\\textbf{Backbone:} IResNet-50/100/150\cite{duta2021improved} w/ Squeeze-and-Excitation blocks~\cite{hu2018squeeze}\strut } \\\hline

CTAI & China & All & \multicolumn{1}{c}{-} &\mline{2cm}{DCFace\cite{kim2023dcface}\\GANDiff\cite{melzi2023gandiffface}} & \mline{\dataw}{They cleaned the data using an IResNet with Squeeze-and-Excitation blocks. They also used DBSCAN to segregate intra-class noise and removed the IDs that were far from the ID class center.} & \mline{\trainw}{\strut They used two models trained with different losses and occlusion augmentation. The final score was obtained by combining the outputs. \\ \textbf{Loss:} AdaFace \& CosFace~\cite{wang2018cosface}\\\textbf{Backbone:} IResNet-100\cite{duta2021improved}\strut } \\\hline

Idiap-SynthDistill & Switzerland & \mline{1cm}{1.2\\2.2} & \ref{fig:Idiap} & Novel & \mline{\dataw}{They used an end-to-end method that generates synthetic images while training the FR model. The base data was generated with a StyleGAN2~\cite{karras2019style} model and during training, they dynamically generated more images based on the training loss.} & \mline{\trainw}{\strut They used a pretrained model to train by model distillation a new model using only the synthetic data they generated.\\ \textbf{Loss:} SynthDistill~\cite{otroshishahreza2024knowledge}\\\textbf{Backbone:} IResNet-101\cite{duta2021improved}\strut } \\ \hline

INESC-IGD & \mline{1cm}{Portugal\\Germany} & All & \ref{fig:inesc} & \mline{2cm}{DCFace\cite{kim2023dcface}\\IDiff~\cite{boutros2023idiff}} & \mline{\dataw}{They labeled the images from the public databases with the ethnicity and balanced their final data.} & \mline{\trainw}{\strut They trained two models using the same loss and applying RandAug and occlusion augmentation. The final score was obtained by combining the outputs from both models. \\ \textbf{Loss:} ElasticCosFac-Plus~\cite{boutros2022elasticface}\\\textbf{Backbone:} ResNet-100~\cite{he2016deep}\strut }\\ \hline

UNICA-IGD-LSI & \mline{1cm}{Italy\\Germany\\Slovenia} & All & \ref{fig:Unica} & DCFace\cite{kim2023dcface} & \mline{\dataw}{They used the public data from DCFace and generated data with the IDiff-Face Framework and ExFaceGAN~\cite{boutros2023exfacegan}.} & \mline{\trainw}{\strut They took the similarity mean between real and synthetic samples and added it to the loss value. \\ \textbf{Loss:} CosFace\cite{wang2018cosface}\\\textbf{Backbone:} ResNet-101\cite{he2016deep}\strut } \\ \hline

SRCN\_AIVL & China & 1.1 & \ref{fig:SRCN} & \mline{2cm}{DCFace\cite{kim2023dcface}\\IDiff\cite{boutros2023idiff}} & \mline{\dataw}{They balanced DCFace in ethnicity and trained IDiff-Face to be balanced as well.} & \mline{\trainw}{\strut They trained two models and for the inference, they ingested one with the original image and the other with the flipped image. Then they merged the two resulting vectors and used the resulting vector to get the distance scores.\\ \textbf{Loss:} AdaFace\cite{kim2022adaface}\\\textbf{Backbone:} ResNet-101\cite{he2016deep}\strut } \\ \hline

CBSR-Samsung & China & \mline{1cm}{1.3\\ 2.3} & \ref{fig:CBSR} & DCFace\cite{kim2023dcface} & \mline{\dataw}{They de-overlaped DCFace from CASIA.} & \mline{\trainw}{\strut They created a validation dataset using a subset of the DCFace, adding masks and sunglasses and positioning vertical bars in the images. \\ \textbf{Loss:} AdaFace\cite{kim2022adaface}\\\textbf{Backbone:} ResNet-100\cite{he2016deep}\strut } \\ \hline

BOVIFOCR-UFPR & Brazil & \mline{1cm}{1.2\\ 2.1} & \ref{fig:BOVIFOCR} & DCFace\cite{kim2023dcface} & \multicolumn{1}{c}{-} &\mline{\trainw}{\strut They used RandAugment, random erasing and random flip as augmentation techniques. \\ \textbf{Loss:} ArcFace\cite{deng2019arcface}\\\textbf{Backbone:} ResNet-100\cite{he2016deep}\strut } \\ \bottomrule
\end{tabular}}
\end{table}
\end{landscape}

\restoregeometry

\begin{figure*}[t]
    \centering
    \includegraphics[width=\linewidth]{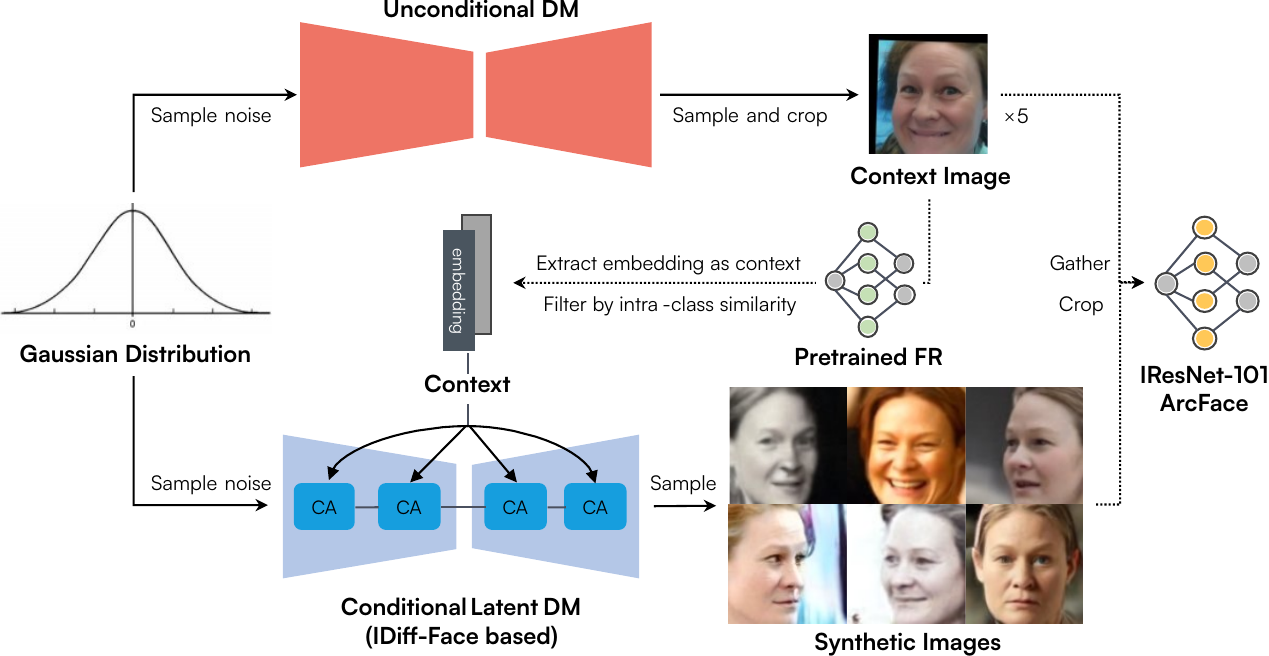}
    \caption{Framework proposed by the ADMIS team.}
    \label{fig:Admis}
\end{figure*}
\paragraph*{\textbf{ADMIS (All sub-tasks)}}
This team comprises members from Fudan University and Tencent Youtu Lab, China. They used a Latent Diffusion Model (LDM)~\cite{rombach2022high} based on IDiff-Face~\cite{boutros2023idiff} to synthesize faces. While IDiff-Face uses a noise embedding sampled from a Gaussian distribution to serve as the LDM's, the ADMIS team uses identity embeddings as contexts, extracted from faces by a pretrained ElasticFace~\cite{boutros2022elasticface} model with an IResNet-101~\cite{duta2021improved} backbone. They trained the LDM with the CASIA-WebFace~\cite{yi2014learning} database. As the LDM takes the ID embeddings as context, they considered an unconditional Denoising Diffusion Probabilistic Model (DDPM) trained on the FFHQ database~\cite{karras2019style} as a context generator. Specifically, they used the DDPM to generate $400K$ faces with arbitrary identities, known as context faces. They exploited the same ElasticFace model to extract the embeddings from the context faces. To encourage the quality and the distinctiveness of identity from later LDM-generated faces, they filtered the embeddings by setting a minimum cosine similarity threshold of $0.3$ between arbitrary pairs of embeddings. This yields $\approx$$30K$ embeddings with discriminative identities. Furthermore, they accelerated the sampling process of the LDM by Denoising Diffusion Implicit Models (DDIM)~\cite{song2021denoising}. For the training of the FR model, they generated $49$ images for each context. They adopted the ID oversampling strategy from DCFace~\cite{kim2023dcface} and performed it five times for each ID to enhance consistency. As a result, $10K$ contexts were used for Sub-Tasks 1.1 and 2.1, while $30K$ for Sub-Tasks 1.2 and 2.2. For Sub-Tasks 1.3 and 2.3, they expanded Sub-Tasks 1.1 and 2.1 with the CASIA-WebFace database. Both the baseline and proposed FR models used IResNet-101 architectures. They applied the ArcFace~\cite{deng2019arcface} loss with a batch size of $64$ and an initial learning rate of $0.1$ for $40$ epochs. The learning rate was divided by $10$ at epochs $22$, $28$, and $32$. They also used random cropping augmentation during training. Their proposed architecture is described in Figure~\ref{fig:Admis}.

\noindent Code: \footnotesize\href{https://github.com/zzzweakman/CVPR24_FRCSyn_ADMIS}{https://github.com/zzzweakman/CVPR24\_FRCSyn\_ADMIS}\normalsize

\begin{figure*}[t]
    \centering
    \includegraphics[width=\linewidth]{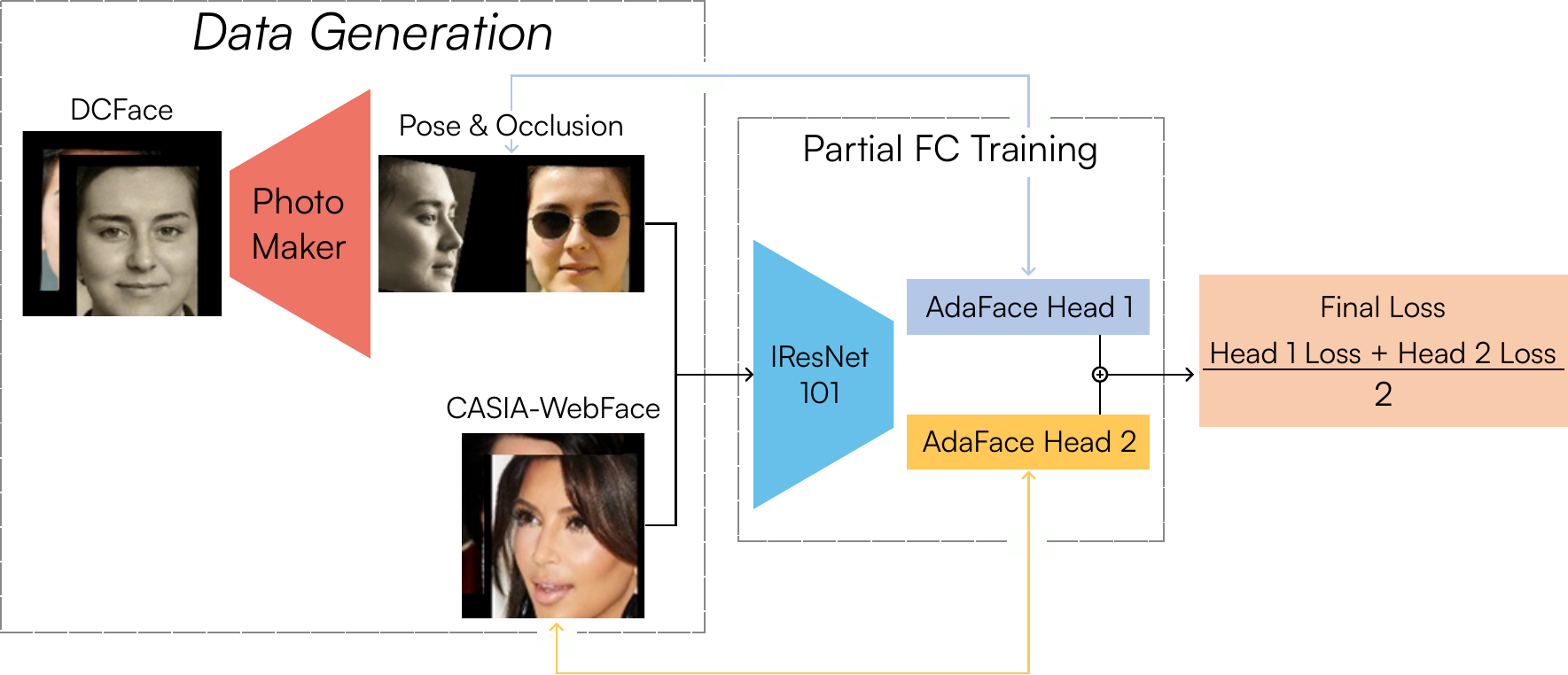}
    \caption{Framework proposed by the OPDAI team.}
    \label{fig:OPDAI}
\end{figure*}

\paragraph*{\textbf{OPDAI (All sub-tasks)}}
This team comprises members from the Interactive Entertainment Group of Netease Inc., China. They initially used the DCFace~\cite{kim2023dcface} database, generating then $10$ more face images for each ID with large pose variations and occlusions using Photomaker~\cite{li2024photomaker}. Given a few input ID images, PhotoMaker can generate diverse personalized ID photos based on a text prompt while preserving the identity information from the input image. They randomly replaced these images in the original DCFace data to ensure that the total number of samples meets the requirement of $500K$. During the Photomaker inference, they adopted a batch size of $1$ and used random prompts including age, pose, and image quality to ensure the diversity of the generated samples. For Sub-Tasks 1.2 and 2.2, they combined this data with the $1.2M$ version of DCFace, while for Sub-Tasks 1.3 and 2.3, it was merged with CASIA-WebFace~\cite{yi2014learning}. For Sub-Tasks 1.2, 1.3, 2.2, and 2.3 they did not merge nor denoise samples from different databases, following the Partial FC approach~\cite{an2021partial}, which consists in a sparse variant of model parallel architecture for training FR models. Regarding the FR model, they obtained the loss of different databases in independent AdaFace~\cite{kim2022adaface} heads, calculating the final loss as the average of the multiple heads. Both baseline and proposed models are based on IResNet-100~\cite{duta2021improved} architectures, with horizontal flipping. Their proposed architecture is described in Figure~\ref{fig:OPDAI}.

\noindent Code: \footnotesize\href{https://github.com/mightycatty/frcsyn_cvpr2024.git}{https://github.com/mightycatty/frcsyn\_cvpr2024.git}\normalsize

\begin{figure*}[t]
    \centering
    \includegraphics[width=\linewidth]{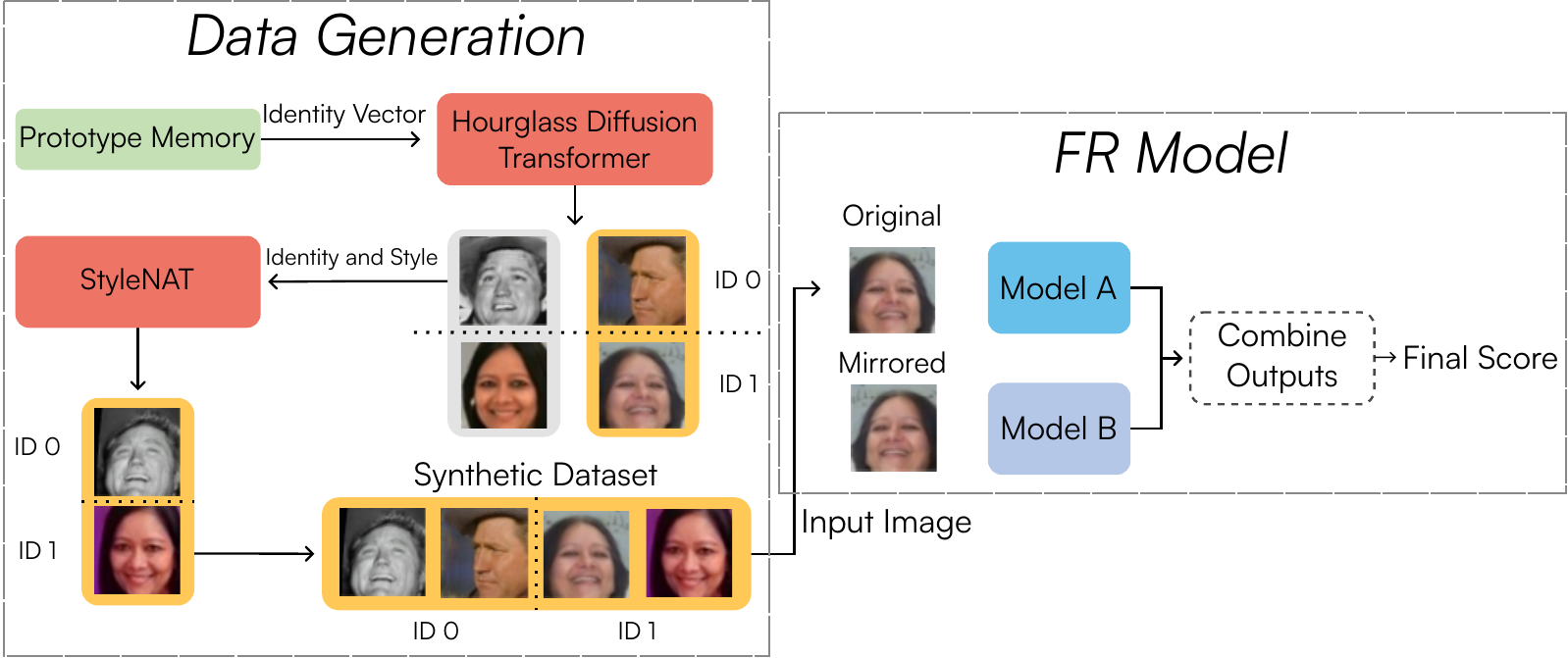}
    \caption{Framework proposed by the ID R\&D team.}
    \label{fig:IDRnD}
\end{figure*}

\paragraph*{\textbf{ID R\&D (All sub-tasks)}}
This team comprises members from ID R\&D Inc, USA. To generate the synthetic data, they used two models trained on WebFace42M~\cite{zhu2021webface260m}. The first model was based on Hourglass Diffusion Transformers (HDT)~\cite{crowson2024scalable}, which combines the scalability of Transformer architectures with the efficiency of convolutional U-Nets~\cite{ronneberger2015u}. It was trained focusing on conditional flow matching~\cite{tong2024improving} and following the classifier-free guidance approach~\cite{ho2022classifier}. Identity embeddings were used directly, whereas style embeddings were processed through a Vector Quantised-Variational AutoEncoder (VQVAE)~\cite{oord2017neural}. Specifically, for head position, $32$ embeddings were allocated for VQVAE processing, while age and facial expression attributes were represented with $8$ embeddings each. During inference, combinations of these embeddings were randomly selected. The second model used to generate synthetic data was based on StyleNAT~\cite{walton2022stylenat}, enhanced with a FR model. This network was trained using auxiliary sources of supervision: a pre-trained FR network with Prototype Memory~\cite{smirnov2022prototype} (for identity supervision) and a pre-trained face attribute classification network (for style supervision). To create their synthetic data, they used classifier weights of the trained Prototype Memory to get $50K$ identity embeddings, of which $20K$ were randomly selected and $30K$ were uniformly sampled from the $1K$ clusters obtained using \textit{k}-means, to get demographic diversity. For each identity, they generated $5$ face images using each of the two generative models. In the first stage, identity embeddings were used by the HDT model to get $10$ images for each identity. Then $5$ of these images were included in the training dataset. Identity and style embeddings were taken from the remaining $5$ images and used as a condition to generate $5$ different images with the StyleNAT model. These images were also put into the training dataset. Regarding the FR model, they trained an IResNet-200~\cite{duta2021improved} with UniFace~\cite{zhou2023uniface} loss for $28$ epochs. One network was trained with color, geometric augmentations, and FaceMix-B~\cite{garaev2023facemixa}, and the other one using only random horizontal flipping. These two networks were combined in an ``ensemble", where the first one received the original image, and the second one a mirrored copy. They used the same model for Sub-Tasks 1.1, 1.2, 2.1 and 2.2. For Sub-Tasks 1.3 and 2.3, they combined the synthetic data with the CASIA-WebFace~\cite{yi2014learning} data, training two models, one on the mixed data, and the other on the CASIA-WebFace. Their proposed architecture is described in Figure~\ref{fig:IDRnD}.

\begin{figure*}[t]
    \centering
    \includegraphics[width=\linewidth]{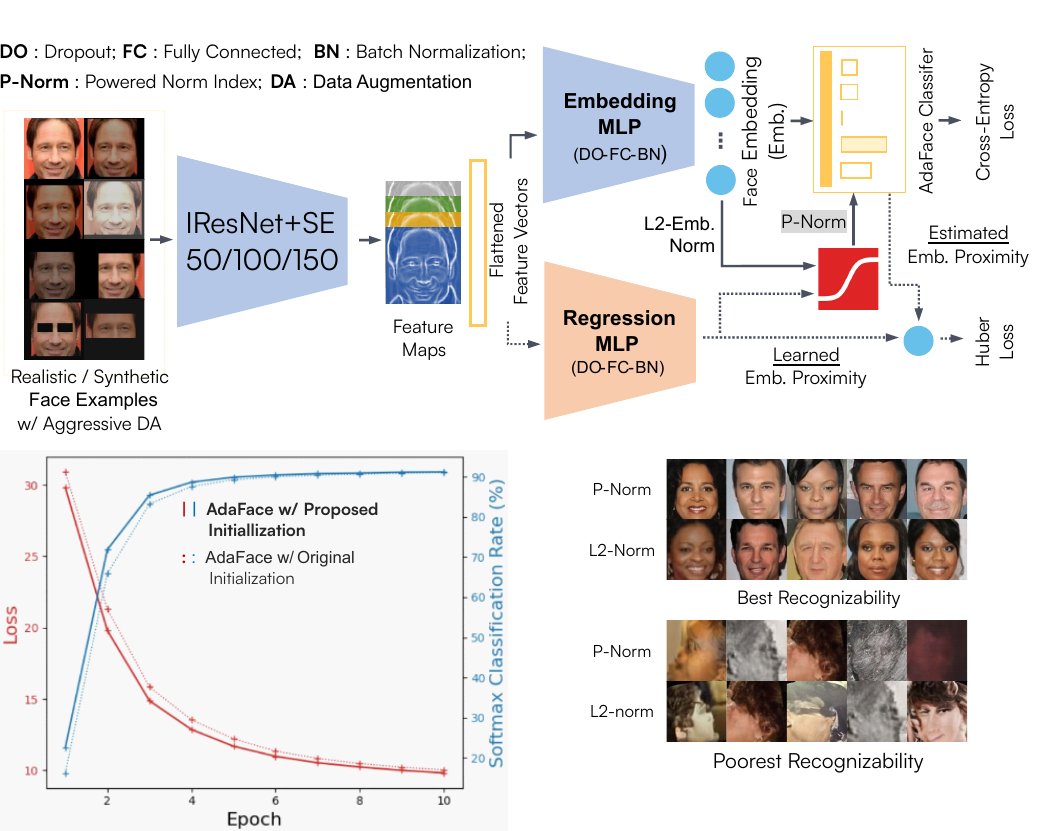}
    \caption{Framework proposed by the K-IBS-DS team.}
    \label{fig:KIBS}
\end{figure*}

\paragraph*{\textbf{K-IBS-DS (All sub-tasks)}}
This team comprises members from the Korea Advanced Institute of Science \& Technology and the Institute for Basic Science, South Korea. They used DCFace~\cite{kim2023dcface} with $500K$ and $1.2M$ face images (depending on the sub-task). Regarding the FR model, they used several IResNet~\cite{duta2021improved} models of $50$, $100$, and $152$ layers with Squeeze-and-Excitation blocks~\cite{hu2018squeeze}, which are architectural components designed to enhance the representational power of convolutional neural networks by dynamically adjusting channel-wise features. Inspired by SlackedFace~\cite{low2023slackedface}, they made two modifications to enhance the AdaFace~\cite{kim2022adaface} FR classifier: \textit{i)} they used renormalized uniform initialization as a more reliable weight initialization for uniformity across identity prototypes in the unit sphere and \textit{ii)} replaced the L2-norm with the powered-norm (p-norm), or face recognizability index from~\cite{low2023slackedface}, which integrates the L2-norm with the learned embedding proximity. The training stage was in line with~\cite{kim2023dcface} and~\cite{bae2023digiface}, including optimizer, learning rate, etc. For Sub-Tasks 1.3 and 2.3, the first $10K$ subjects of the CASIA-WebFace~\cite{yi2014learning} were assigned for training, and the remaining ones for performance validation using random pairs with challenging conditions (identified based on the poorest L2-norm values~\cite{melzi2023gandiffface}). The final score was obtained by aggregating the comparison scores of the different IResNet models, along with the horizontally flipped instances through score fusion. All training and test sets were realigned using RetinaFace~\cite{deng2020retinaface}, followed by a similarity transformation. For all sub-tasks, aggressive data augmentations were applied, including random horizontal flipping, photometric operations, cropping, resizing, and the addition of sunglasses and masks. Their proposed architecture is described in Figure~\ref{fig:KIBS}.
\noindent Code: \footnotesize\href{https://github.com/kalebmes/cvpr_frcsyn}{https://github.com/kalebmes/cvpr\_frcsyn}\normalsize

\paragraph*{\textbf{CTAI (All sub-tasks)}}
This team comprises members from China Telecom AI, China. By analyzing popular synthetic data, they found that intra-class and inter-class noise was widely present. Data cleaning can effectively remove the bad examples of synthetic data and retain important images from a large amount of synthetic data. In order to select the optimal synthetic data, they first trained an IResNet-100~\cite{duta2021improved} model with Squeeze-and-Excitation blocks~\cite{hu2018squeeze} using CASIA-WebFace~\cite{yi2014learning} to extract features of synthetic images from DCFace~\cite{kim2023dcface}, GANDiffFace~\cite{melzi2023gandiffface}, and DigiFace-1M~\cite{bae2023digiface}. Subsequently, they used the DBSCAN clustering method to segregate intra-class noise and removed IDs with a class center feature cosine similarity greater than $0.5$. Finally, they used the cleaned synthetic data merged with CASIA-WebFace to finetune the IResNet-100 for a second data refinement. From the final refined synthetic dataset, they sampled $500K$ face images while retaining as many IDs as possible to build their synthetic training set. Regarding the FR model, in particular Sub-Task 2.3 in which they achieved their highest position among all sub-tasks, they trained IResNet-100 with AdaFace loss~\cite{kim2022adaface} (A1) and CosFace loss~\cite{wang2018cosface} (A2) with mask and occlusion augmentation on CASIA-WebFace and the refined synthetic data. They used an ensemble of A1, A2, and a model trained with only synthetic data. Furthermore, data augmentation was considered to enhance all features. 

\noindent Code: \footnotesize\href{https://github.com/liuhao-lh/FRCSyn-Challenge}{https://github.com/liuhao-lh/FRCSyn-Challenge}\normalsize

\begin{figure*}[t]
    \centering
    \includegraphics[width=0.85\linewidth]{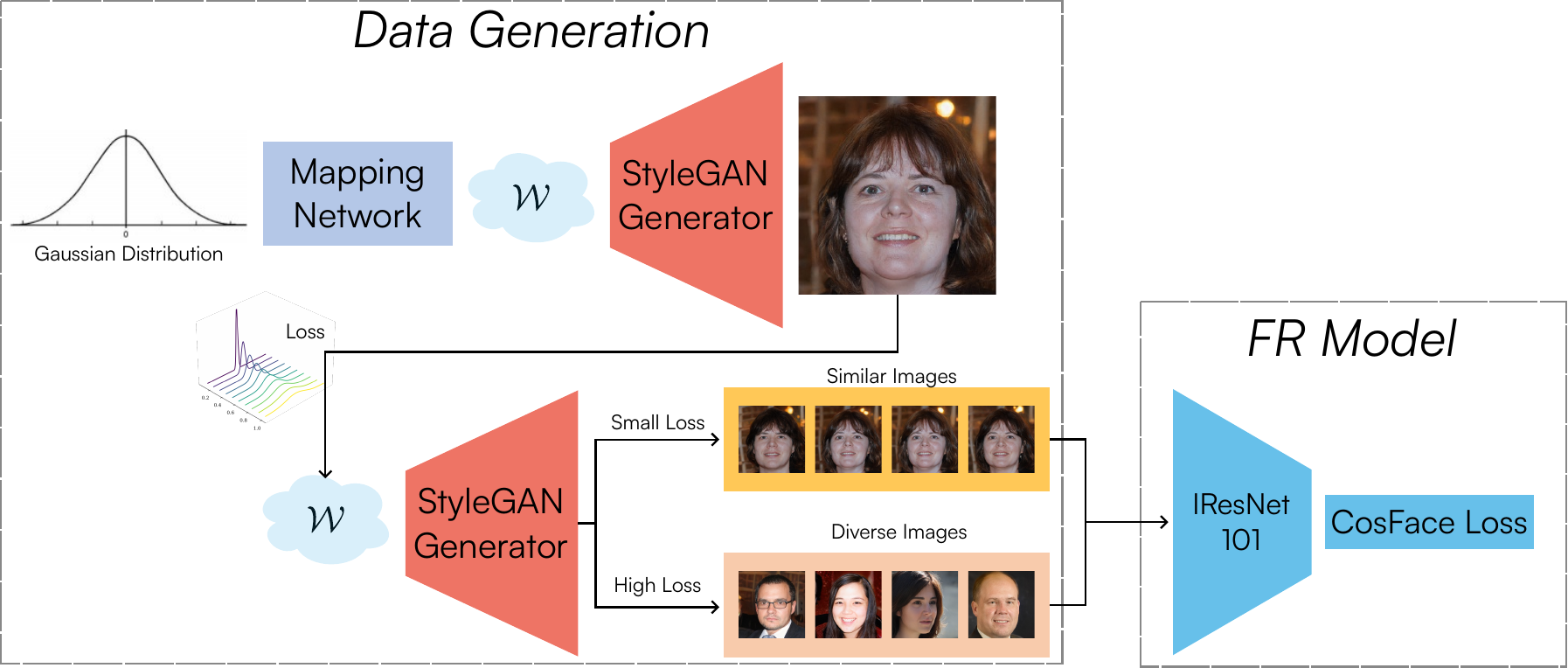}
    \caption{Framework proposed by the Idiap-SynthDistill team.}
    \label{fig:Idiap}
\end{figure*}

\paragraph*{\textbf{Idiap-SynthDistill (Sub-Tasks 1.2 and 2.2)}}
This team comprises members from the Idiap Research Institute, EPFL, and Université de Lausanne, Switzerland. The proposed method was based on SynthDistill~\cite{otroshishahreza2024knowledge}, which is an end-to-end approach, generating synthetic images and training the FR model in the same training loop. Instead of using the pre-trained model in a separate step, they directly used it in the training loop for supervision, while a new student FR model was trained fully using synthetic data generated from a StyleGAN model~\cite{karras2019style}. For generating synthetic images, they trained StyleGAN2~\cite{karras2020analyzing} with the CASIA-WebFace database~\cite{yi2014learning} and then dynamically generated 20$M$ synthetic images during training based on the training loss. For the dynamic image generation, they used the training loss from every iteration as feedback to find the most difficult synthetic image in each batch and then re-sampled a new batch of synthetic images in the intermediate latent space $\mathcal{W}$ of StyleGAN near the latent vector of the most difficult sample. If the loss value was high, they re-sampled with a relatively small standard deviation around the difficult sample and generated similar images, but if the loss value was small they re-sampled with a higher standard deviation, generating images with more variations. Throughout the process, the generated images were resized to $112 \times 112$ before being fed to the FR models. They used a pre-trained FR model with IResNet-101~\cite{duta2021improved} architecture trained with CosFace~\cite{wang2018cosface} loss on a subset of the WebFace260M database~\cite{zhu2021webface260m} and trained a new model as a student network with the same architecture using synthetic data with their dynamic synthetic image generation approach. They used the Adam optimizer with an initial learning rate of $0.001$ and trained their student model with the same loss function as in~\cite{otroshishahreza2024knowledge}. For thresholding, a subset of DCFace~\cite{kim2023dcface} was used to determine the optimal threshold for maximizing verification accuracy, using a 10-fold cross-validation approach based on a random selection of identities and comparison pairs. Their proposed architecture is described in Figure~\ref{fig:Idiap}.
\begin{figure*}[t]
    \centering
    \includegraphics[width=0.9\linewidth]{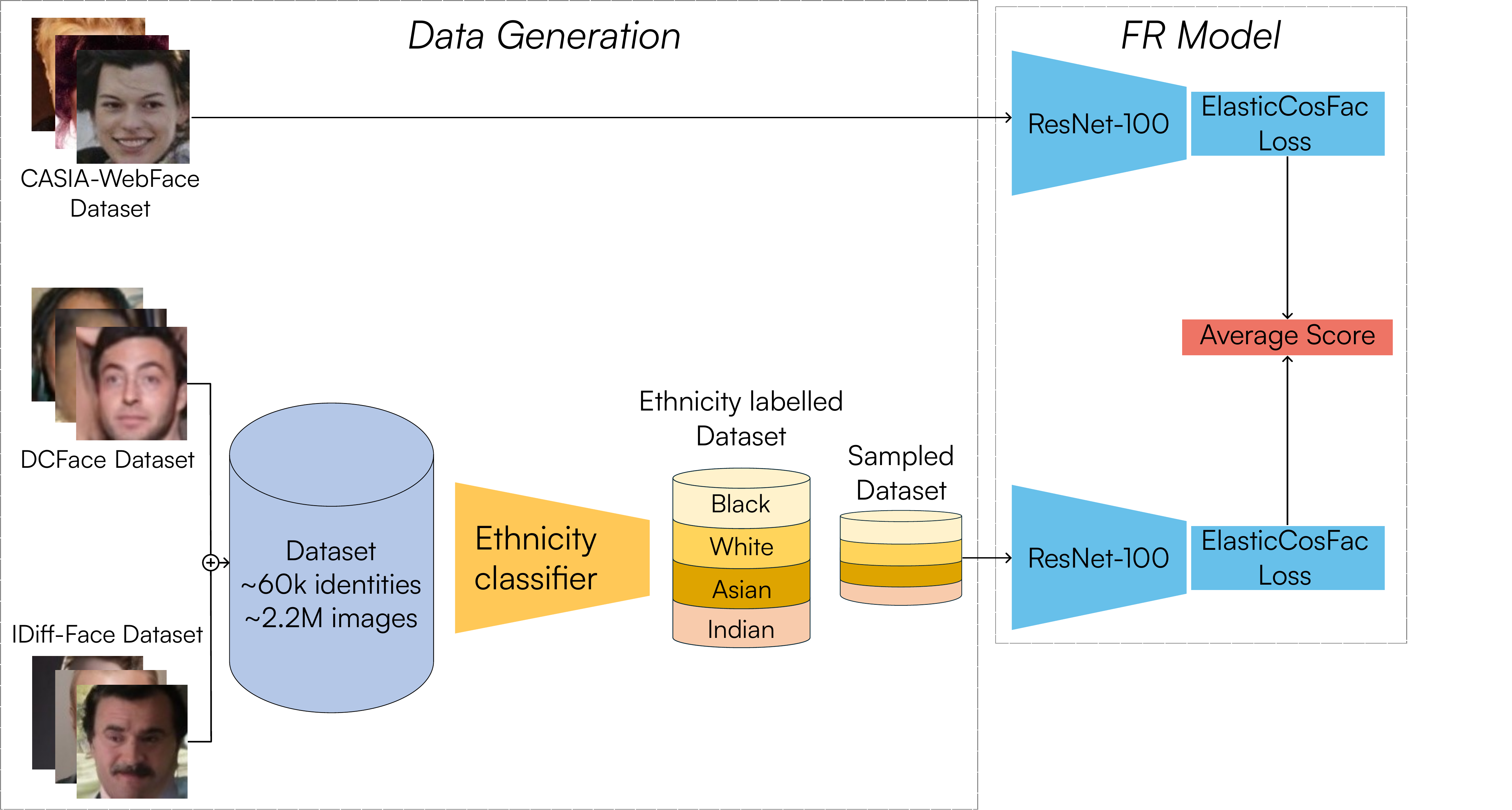}
    \caption{Architecture proposed by the INESC-IGD team.}
    \label{fig:inesc}
\end{figure*}

\noindent Code: \footnotesize\href{https://gitlab.idiap.ch/bob/bob.paper.ijcb2023_synthdistill}{https://gitlab.idiap.ch/bob/bob.paper.ijcb2023\_synthdistill}\normalsize

\paragraph*{\textbf{INESC-IGD (All sub-tasks)}}
This team comprises members from INESC TEC and Universidade do Porto, Portugal, and Fraunhofer IGD, Germany. For the training dataset, they merged DCFace~\cite{kim2023dcface}, IDiff-Face Uniform, and IDiff-Face Two-stage~\cite{boutros2023idiff} databases and then labeled the data with ethnicity labels using a similar approach to~\cite{neto2023compressed}. For Sub-Tasks 1.1 and 2.1, they created a synthetic training dataset containing $500K$ face images by sampling $7K$ balanced identities, in terms of ethnicity labels. For Sub-Tasks 1.2 and 2.2, they created a synthetic training dataset containing $2.1M$ face images by sampling $50K$ identities from the training datasets. For Sub-Tasks 1.3 and 2.3, two instances of ResNet-100~\cite{he2016deep} were trained, one on CASIA-WebFace~\cite{yi2014learning} and the other on a subset of synthetic datasets (\textit{e.g.,} $400K$ images of $9K$ identities). For all sub-tasks, they trained a ResNet-100 with ElasticCosFac-Plus loss~\cite{boutros2022elasticface} using the settings presented in~\cite{boutros2022elasticface}. During the testing phase of Sub-Tasks 1.3 and 2.3, feature embeddings were obtained from trained models and the weighted sum of $0.5$ score-level fusion was used. During the FR training of all sub-tasks, the training datasets were augmented using RandAug and occluded augmentation~\cite{neto2022ocfr} with probabilities of $0.4$. The occluded augmentation followed protocol 4 proposed in \cite{neto2022ocfr}, leading to occlusions on the eyes, lower face, upper face, or a combination of the eyes occlusion with the others. Occluded augmentations boosted the performance, as synthetic data has a lower frequency of natural occlusions such as beard and makeup \cite{neto2024massively}. Their proposed architecture is described in Figure~\ref{fig:inesc}.

\noindent Code: \footnotesize\href{https://github.com/NetoPedro/Equilibrium-Face-Recognition}{https://github.com/NetoPedro/Equilibrium-Face-Recognition}\normalsize
\begin{figure*}[t]
    \centering
    \includegraphics[width=\linewidth]{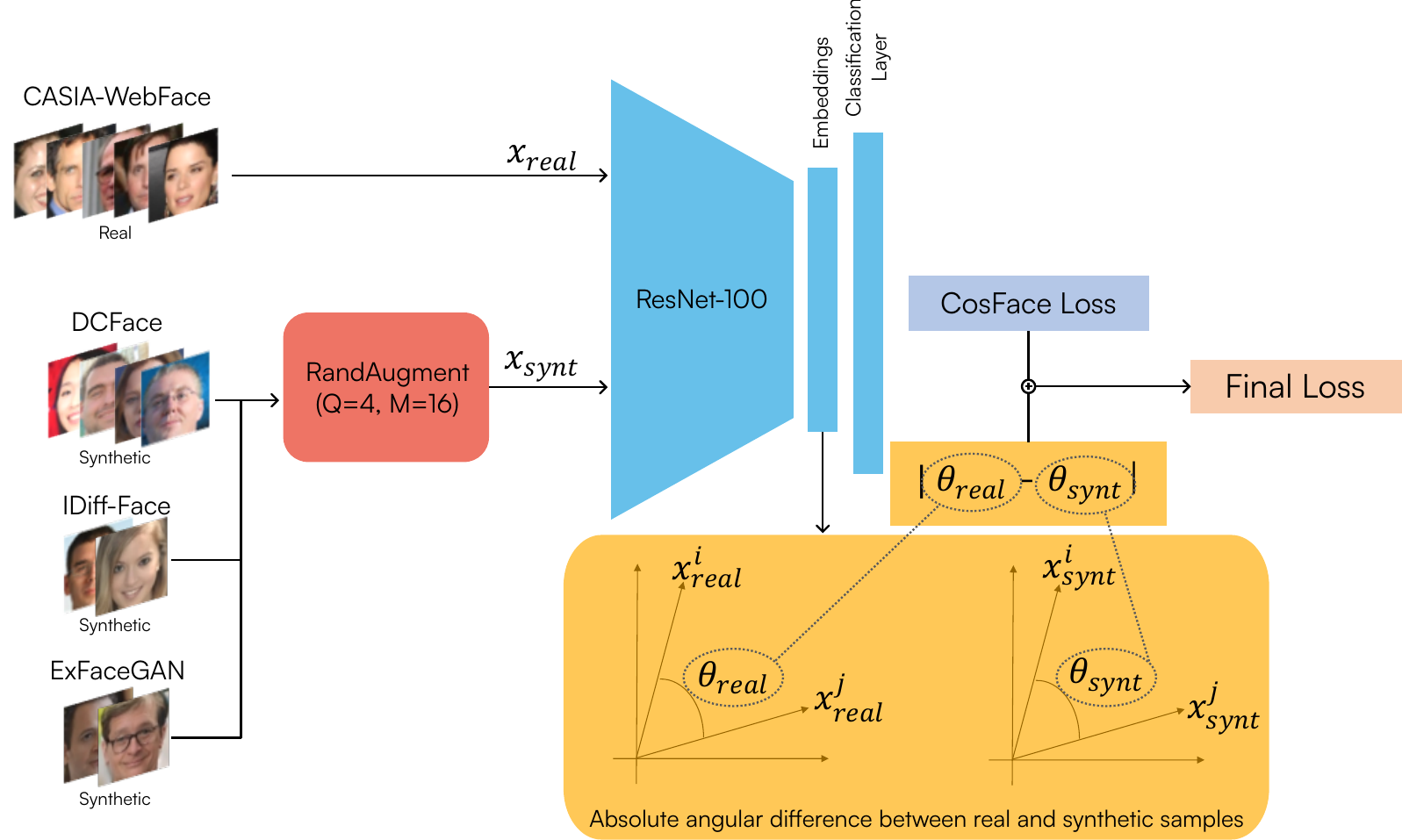}
    \caption{Framework proposed by the UNICA-IGD-LSI team.}
    \label{fig:Unica}
\end{figure*}
\paragraph*{\textbf{UNICA-IGD-LSI (All sub-tasks)}}
This team comprises members from Fraunhofer IGD, Germany, University of Cagliari, Italy, and University of Ljubljana, Slovenia. They used the DCFace~\cite{kim2023dcface} synthetic database as it led to remarkable performance gains under well-known evaluation benchmarks for face verification, while combined with real data~\cite{atzori2024if}. Also, they considered synthetic data generated with IDiff-Face~\cite{boutros2023idiff} and ExFaceGAN~\cite{boutros2023exfacegan} in Sub-Tasks 1.1, 1.2, 2.1, and 2.2. The ExFaceGAN data was generated using an identity disentanglement approach on pretrained GAN-Control~\cite{shoshan2021gan}. Regarding the FR model, they trained a ResNet-100~\cite{he2016deep} network using CosFace loss~\cite{wang2018cosface} with a margin penalty of 0.35 and a scale term of 64. The similarity mean difference between real-only and synthetic-only samples ($|\theta_{real} - \theta_{synt}|$) was scaled and added to the loss value. They trained the FR model for $40$ epochs with a batch size of $512$ and an initial learning rate of $0.1$, which was divided by $10$ after $10$, $22$, $30$, and $40$ epochs. During the training phase, the synthetic samples were augmented using RandAugment with $4$ operations and a magnitude of 16, following~\cite{boutros2023unsupervised, atzori2024if}. For Sub-Tasks 1.3 and 2.3, the selected synthetic dataset was combined with CASIA-Webface~\cite{yi2014learning}, obtaining a total of $1M$ images from $20,572$ identities. Their proposed architecture is described in Figure~\ref{fig:Unica}.

\noindent Code: \footnotesize\href{https://github.com/atzoriandrea/FRCSyn2}{https://github.com/atzoriandrea/FRCSyn2}\normalsize

\begin{figure*}[t]
    \centering
    \includegraphics[width=\linewidth]{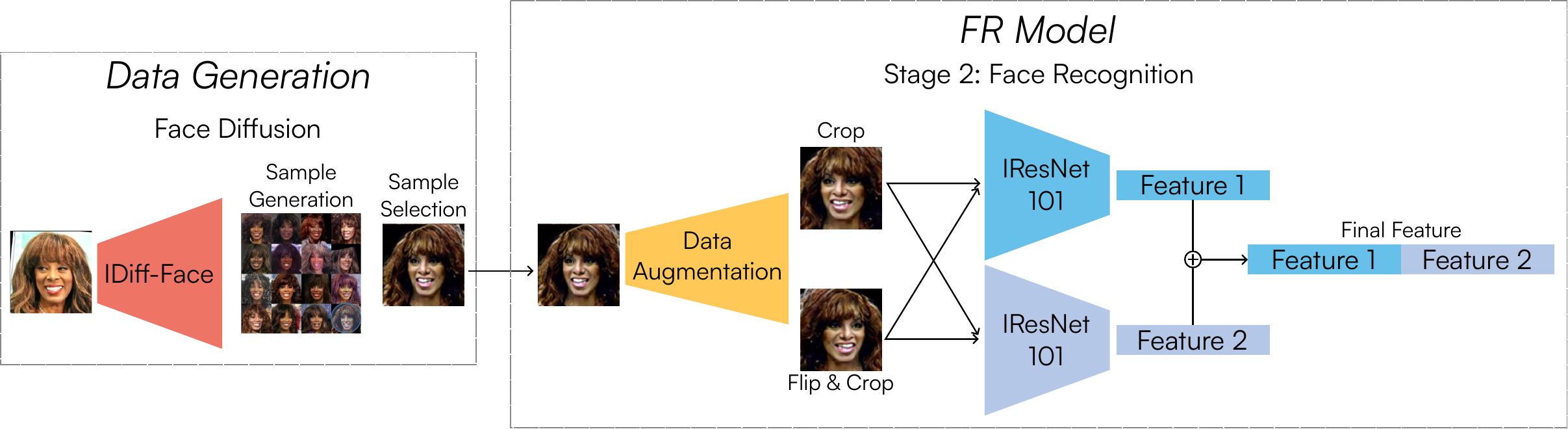}
    \caption{Framework proposed by the SRCN\_AIVL team.}
    \label{fig:SRCN}
\end{figure*}

\paragraph*{\textbf{SRCN\_AIVL (Sub-Task 1.1)}}
This team comprises members from Samsung Electronics (China) R\&D Centre, University of Science and Technology, IIE, CAS, and MAIS, CASIA, China. They selected $400K$ samples from the DCFace~\cite{kim2023dcface} database and labeled the ethnicity of each subject, as the racial distribution gap may lead to bad performance in testing. Based on this approach, they trained IDiff-Face~\cite{boutros2023idiff} with CASIA-WebFace~\cite{yi2014learning} database generating $100K$ synthetic face images of specific races. Regarding the FR system, they used two custom ResNet-101~\cite{he2016deep} trained with AdaFace loss~\cite{kim2022adaface} function. The models were trained for $60$ epochs with an initial learning rate of $0.1$ and a batch size of $512$, which was adjusted at predefined milestones. Their training data underwent further preprocessing, including padding crop augmentation, low-resolution augmentation, photometric augmentation, random grayscale, and normalization. The threshold was determined by the 10-fold optimal threshold in the validation set. For the inference, data preprocessing involved an MTCNN~\cite{zhang2016joint} for the alignment and resizing all data to $112 \times 112$. After cropping and alignment, they fed the image and the flipped image into the two models. Finally, after obtaining the two feature embeddings, they combined them and performed the similarity calculation with these embeddings. Their proposed architecture is described in Figure~\ref{fig:SRCN}.

\noindent Code: \footnotesize\href{https://github.com/Value-Jack/2nd-Edition-FRCSyn}{https://github.com/Value-Jack/2nd-Edition-FRCSyn}\normalsize

\begin{figure*}[t]
    \centering
    \includegraphics[width=\linewidth]{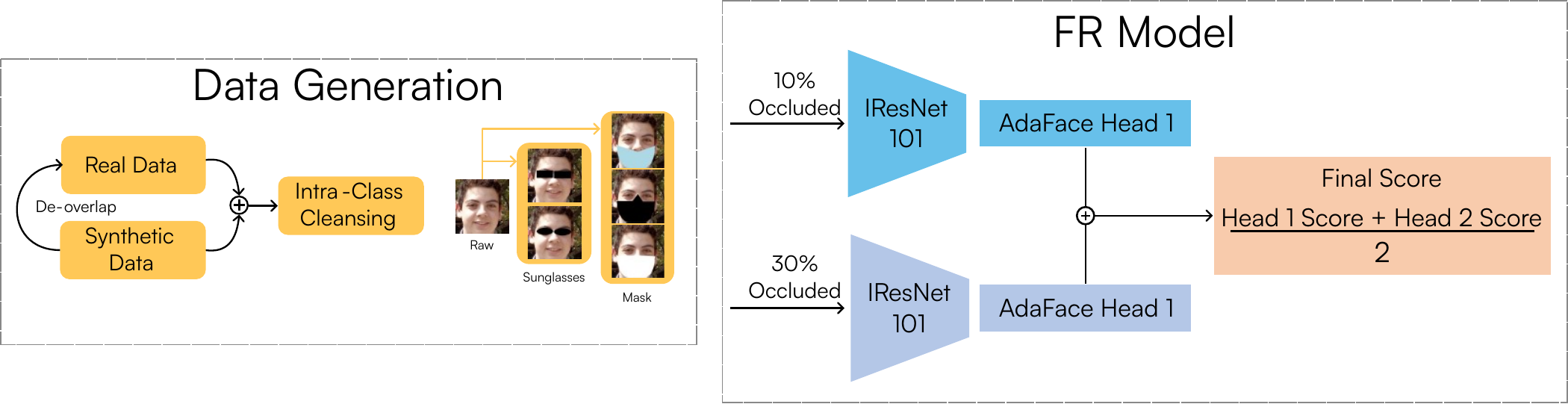}
    \caption{Framework proposed by the CBSR-Samsung team.}
    \label{fig:CBSR}
\end{figure*}
\paragraph*{\textbf{CBSR-Samsung (Sub-Tasks 1.3 and 2.3)}}
This team comprises members from Samsung Electronics (China) R\&D Centre, IIE, CAS, and MAIS, CASIA, China. They first trained a FR model using CASIA-WebFace~\cite{yi2014learning}. Then, they used it to de-overlap DCFace~\cite{kim2023dcface} from CASIA, as DCFace was trained using that real database. For the synthetic dataset, they compared the performance of models trained with three synthetic databases, including GANDiffFace~\cite{melzi2023gandiffface}, DCFace, and IDiff-Face~\cite{boutros2023idiff}, and finally selected DCFace as the only synthetic training set. They created a validation dataset including three subsets for three different testing scenarios: \textit{i)} random sample pairs from DCFace, simulating age variability and demographic groups as in AgeDB~\cite{moschoglou2017agedb} and BUPT-Balanced~\cite{wang2020mitigating} databases, respectively; \textit{ii)} randomly positioned vertical bar masks to the images to simulate the self-occlusion due to as considered in CFP-FP database~\cite{sengupta2016frontal}; and \textit{iii)} add mask and sunglasses to the images by detecting the landmarks via FaceX-Zoo~\cite{wang2021facex}, simulating the ROF database~\cite{erakιn2021recognizing}. This is done following~\cite{ngan2020ongoing}. All validation subsets consist of $6K$ positive pairs and $6K$ negative pairs. Finally, they concatenated these subsets as the validation set. Subsequently, they conducted an intra-class clustering for all datasets using DBSCAN ($0.3$ threshold) and removed the samples that were separated from the class center. Regarding the FR model, they merged the refined datasets and trained IResNet-100~\cite{duta2021improved} with AdaFace loss~\cite{kim2022adaface}. In addition, they adopted two augmentation strategies, \textit{i.e.,} photometric augmentation and rescaling. After that, they trained two FR models using occlusion augmentation with $10\%$ and $30\%$ probability, respectively. Finally, they submitted the average similarity score of the two models. Their proposed architecture is shown in Figure~\ref{fig:CBSR}.

\begin{figure*}[t]
    \centering
    \includegraphics[width=0.9\linewidth]{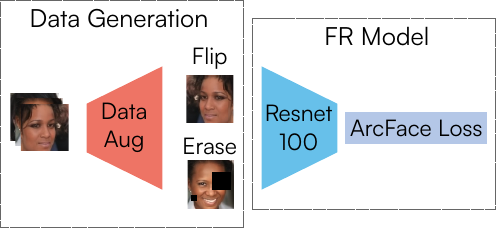}
    \caption{Framework proposed by the BOVIFOCR-UFPR team.}
    \label{fig:BOVIFOCR}
\end{figure*}

\paragraph*{\textbf{BOVIFOCR-UFPR (Sub-Tasks 1.2 and 2.1)}}
This team comprises members from Federal University of Paraná, Federal Institute of Mato Grosso, and unico - idTech, Brazil. They chose DCFace~\cite{kim2023dcface} as the synthetic database and randomly removed $910$ identities with $55$ images per ID to reduce the number to follow the rules. For the FR model, they used a ResNet-100~\cite{he2016deep} as the backbone, trained with the ArcFace~\cite{deng2019arcface} loss function. The images used for training were augmented using a Random Flip with a probability of $0.5$. They also applied random erasing and RandAugment as additional augmentations. To validate their model they subsampled images from DCFace, generating genuine and impostor pairs, and used these pairs to select the best threshold to classify the proposed model output scores. Their proposed architecture is described in Figure~\ref{fig:BOVIFOCR}.

\noindent Code: \footnotesize\href{https://github.com/PedroBVidal/insightface}{https://github.com/PedroBVidal/insightface}\normalsize

\begin{table*}[t!]
\centering
\caption{Results of the teams that ranked among the top-6 in at least one sub-task, ordered by the average rank in all the sub-tasks. For each team, we report the ranking metric and the position across all the sub-tasks. The best result of each sub-task is highlighted in bold. We mark with a `-' if the team did not participate in a sub-task. TO = Trade-off, AVG = average accuracy}
\label{tab:results}
\resizebox{0.8\linewidth}{!}{\begin{tabular}{lcccccc}
\hline
&\multicolumn{3}{|c|}{\textbf{Task 1: Bias Mitigation}} & \multicolumn{3}{c}{\textbf{Task 2: Overall Improvement}} \\ \hline

\textbf{Team} & \textbf{\begin{tabular}[c]{@{}c@{}}Task 1.1\\ TO {[}\%{]}\end{tabular}} & \textbf{\begin{tabular}[c]{@{}c@{}}Task 1.2\\ TO {[}\%{]}\end{tabular}} & \textbf{\begin{tabular}[c]{@{}c@{}}Task 1.3\\ TO {[}\%{]}\end{tabular}} & \textbf{\begin{tabular}[c]{@{}c@{}}Task 2.1\\ AVG {[}\%{]}\end{tabular}} & \textbf{\begin{tabular}[c]{@{}c@{}}Task 2.2\\ AVG {[}\%{]}\end{tabular}} & \textbf{\begin{tabular}[c]{@{}c@{}}Task 2.3\\ AVG {[}\%{]}\end{tabular}} \\ \hline
ADMIS & 94.30 (2) & 95.72 (2) & \textbf{96.50 (1)} & 91.19 (3) & 92.92 (2) & 94.15 (5) \\
OPDAI & 93.75 (4) & 94.12 (3) & 95.96 (4) & \textbf{91.93 (1)} & 92.04 (3) & 95.23 (2) \\
ID R\&D & \textbf{96.73 (1)} & \textbf{96.73 (1)} & 86.73 (8) & 91.86 (2) & 91.86 (4) & 94.05 (6) \\
K-IBS-DS & 92.91 (6) & 93.72 (5) & 96.17 (2) & 91.05 (4) & 91.61 (5) & \textbf{95.42 (1)} \\
CTAI & 93.21 (5) & 93.21 (6) & 95.41 (7) & 90.59 (5) & 90.59 (6) & 94.56 (3) \\
Idiap-SynthDistill & - & 89.70 (9) & - & - & \textbf{93.50 (1)} & - \\
INESC-IGD& 92.28 (7) & 94.05 (4) & 95.65 (5) & 83.16 (8) & 85.40 (7) & 89.43 (8) \\
UNICA-IGD-LSI & 91.89 (8) & 91.89 (7) & 96.00 (3) & 87.80 (7) & 87.80 (8) & 92.79 (7) \\
SRCN\_AIVL & 94.06 (3)& - & - & - & - & - \\
CBSR-Samsung & - & - & 95.57 (6) & - & - & 94.20 (4) \\
BOVIFOCR-UFPR & - & 90.48 (8) & - & 89.97 (6) & - & - \\ \hline
\end{tabular}}
\end{table*}

\begin{table*}[h!]
\caption{Ranking for the three sub-tasks considered in Task 1. For each sub-task, we highlight in bold the best team according to the Trade-Off. TO = Trade-Off, AVG = Average accuracy, SD = Standard Deviation of accuracy, FNMR = False Non-Match Rate, FMR = False Match Rate, AUC = Area Under Curve, GAP = Gap to Real.}
\label{tab:task1}
\centering
\resizebox{0.8\linewidth}{!}{\begin{tabular}{@{}llcccccc@{}}
\toprule
\multicolumn{8}{c}{\textbf{Sub-Task 1.1 (Bias Mitigation): Synthetic Data (Constrained)}} \\ \midrule
Pos. & Team & TO {[}\%{]} & AVG {[}\%{]} & SD {[}\%{]} & FNMR@FMR=1\% & AUC {[}\%{]} & GAP {[}\%{]} \\ \midrule
\textbf{1} & \textbf{ID R\&D} & \textbf{96.73} & \textbf{97.55} & \textbf{0.82} & \textbf{3.17 (1)} & \textbf{99.57 (1)} & \textbf{-5.31} \\
2 & ADMIS & 94.30 & 95.10 & 0.80 & 11.38 (3) & 98.96 (3) & 1.47 \\
3 & SRCN\_AIVL & 94.06 & 95.12 & 1.07 & 10.72 (2) & 98.83 (4) & -0.54 \\
4 & OPDAI & 93.75 & 94.92 & 1.17 & 11.85 (4) & 99.51 (2) & 1.02 \\
5 & CTAI & 93.21 & 94.74 & 1.53 & 14.38 (5) & 98.33 (6) & -0.63 \\
6 & K-IBS-DS & 92.91 & 94.11 & 1.2 & 15.03 (6) & 98.47 (5) & 1.58 \\ \bottomrule
\end{tabular}}

\bigskip

\resizebox{0.8\linewidth}{!}{\begin{tabular}{@{}llcccccc@{}}
\toprule
\multicolumn{8}{c}{\textbf{Sub-Task 1.2 (Bias Mitigation): Synthetic Data   (Unconstrained)}} \\ \midrule
Pos. & Team & TO {[}\%{]} & AVG {[}\%{]} & SD {[}\%{]} & FNMR@FMR=1\% & AUC {[}\%{]} & GAP {[}\%{]} \\ \midrule
\textbf{1} & \textbf{ID R\&D} & \textbf{96.73} & \textbf{97.55} & \textbf{0.82} & \textbf{3.17 (1) } & \textbf{99.57 (2)} & \textbf{-5.31} \\
2 & ADMIS & 95.72 & 96.50 & 0.78 & 6.33 (2) & 99.51 (3) & -0.56 \\
3 & OPDAI & 94.12 & 95.22 & 1.11 & 10.78 (3) & 98.92 (4) & 0.71 \\
4 & INESC-IGD & 94.05 & 95.22 & 1.17 & 11.03 (4) & 98.70 (5) & 1.04 \\
5 & K-IBS-DS & 93.72 & 94.88 & 1.16 & 12.75 (5) & 99.66 (1) & 0.77 \\
6 & CTAI & 93.21 & 94.74 & 1.53 & 14.38 (6) & 98.33 (6) & -0.63 \\ \bottomrule
\end{tabular}}

\bigskip

\resizebox{0.8\linewidth}{!}{\begin{tabular}{@{}llcccccc@{}}
\toprule
\multicolumn{8}{c}{\textbf{Sub-Task 1.3 (Bias Mitigation): Synthetic + Real Data   (Constrained)}} \\ \midrule
Pos. & Team & TO {[}\%{]} & AVG {[}\%{]} & SD {[}\%{]} & FNMR@FMR=1\% & AUC {[}\%{]} & GAP {[}\%{]} \\ \midrule
\textbf{1} & \textbf{ADMIS} & \textbf{96.50} & \textbf{97.25} & \textbf{0.75} & \textbf{3.90 (1)} & \textbf{99.72 (1)} & \textbf{-1.33} \\
2 & K-IBS-DS & 96.17 & 96.92 & 0.75 & 5.88 (4) & 99.54 (2) & -1.37 \\
3 & UNICA-IGD-LSI & 96.00 & 96.70 & 0.70 & 5.90 (5) & 99.49 (3) & -5.33 \\
4 & OPDAI & 95.96 & 96.80 & 0.84 & 4.90 (2)& 99.54 (2) & -0.03 \\
5 & INESC-IGD & 95.65 & 96.33 & 0.67 & 6.15 (6) & 99.18 (5) & -0.12 \\
6 & CBSR-Samsung & 95.57 & 96.54 & 0.97 & 5.00 (3) & 99.41 (4) & -24.43 \\ \bottomrule
\end{tabular}}
\end{table*}

\section{Second FRCSyn-onGoing: Results}\label{sec:results}
Next, we describe in Sections~\ref{subsec:R_T1} and~\ref{subsec:R_T2} the main results achieved in Tasks 1 and 2, respectively. These results are further analyzed in Section~\ref{subsec:R_Dem} focusing on specific demographic groups and individual databases. Finally, we discuss in Section~\ref{subsec:RCMP} common trends among the different teams and compare the results with those obtained in the 1$^\text{st}$ edition. We present in Table~\ref{tab:results} the ranking and key results of the 2$^\text{nd}$ FRCSyn-onGoing.

\begin{table*}[t]
\caption{Comparison between the baseline model (trained exclusively with real data) and the proposed model (trained with synthetic data) for each finalist of Task 1. The performance of each demographic group is represented by its AVG {[}\%{]}.}
\label{tab:task1_baselinevsproposed}
\centering
\large
\resizebox{\linewidth}{!}{
\begin{tabular}{clccccccccccccc}
\hline
\textbf{\begin{tabular}[c]{@{}c@{}}Sub\\ Task\end{tabular}} & \textbf{Team} & \textbf{Model} & \textbf{\begin{tabular}[c]{@{}c@{}}Asian\\ Female\end{tabular}} & \textbf{\begin{tabular}[c]{@{}c@{}}Asian\\ Male\end{tabular}} & \textbf{\begin{tabular}[c]{@{}c@{}}Black\\ Female\end{tabular}} & \textbf{\begin{tabular}[c]{@{}c@{}}Black\\ Male\end{tabular}} & \textbf{\begin{tabular}[c]{@{}c@{}}Indian\\ Female\end{tabular}} & \textbf{\begin{tabular}[c]{@{}c@{}}Indian\\ Male\end{tabular}} & \textbf{\begin{tabular}[c]{@{}c@{}}White\\ Female\end{tabular}} & \textbf{\begin{tabular}[c]{@{}c@{}}White\\ Male\end{tabular}} & \textbf{\begin{tabular}[c]{@{}c@{}}AVG \\{[}\%{]}\end{tabular}} & \textbf{\begin{tabular}[c]{@{}c@{}}STD \\{[}\%{]}\end{tabular}} & \textbf{\begin{tabular}[c]{@{}c@{}}TO \\{[}\%{]}\end{tabular}} & \textbf{\begin{tabular}[c]{@{}c@{}}GAP \\{[}\%{]}\end{tabular}} \\ \hline
\multirow{2}{*}{\begin{tabular}[c]{@{}c@{}}1.1\\ 1.2\end{tabular}} & \multirow{2}{*}{ID R\&D} & \textbf{Proposed} & \textbf{96.80} & \textbf{96.90} & \textbf{97.20} & \textbf{97.80} & \textbf{98.10} & \textbf{99.20} & 96.50 & 97.90 & \textbf{97.55} & \textbf{0.82} & \textbf{96.73} & \multirow{2}{*}{\textbf{-5.31}} \\ \cline{3-14}
 &  & Baseline & 85.70 & 90.00 & 94.10 & 95.20 & 87.10 & 90.80 & \textbf{97.60} & \textbf{98.40} & 92.36 & 4.40 & 87.96 &  \\ \hline
\multirow{2}{*}{1.3} & \multirow{2}{*}{ADMIS} & \textbf{Proposed} & \textbf{96.20} & \textbf{97.30} & \textbf{97.30} & 97.10 & \textbf{96.20} & \textbf{97.40} & \textbf{98.60} & \textbf{97.90} & \textbf{97.25} & \textbf{0.75} & \textbf{96.50} & \multirow{2}{*}{-1.33} \\ \cline{3-14}
 &  & Baseline & 92.90 & 94.60 & 97.00 & \textbf{98.00} & 93.70 & 96.90 & 97.10 & 97.50 & 95.96 & 1.81 & 94.15 &  \\ \hline
\end{tabular}
}
\end{table*}

\subsection{Task 1: Bias Mitigation}\label{subsec:R_T1}
Table~\ref{tab:task1} shows the results achieved by participants in Task 1, focused on demographic bias mitigation. Teams are ranked by descending order of TO, which tends to correlate to the ascending order of SD (\textit{i.e.,} from less to more biased FR systems). Notably, the winner of Sub-Tasks 1.1 and 1.2, ID R\&D (96.73\% TO), demonstrates a significant negative GAP value (-5.31\%), showing a higher performance when training the FR system with synthetic data compared to real data (\textit{i.e.,} CASIA-WebFace~\cite{yi2014learning}). Furthermore, in Sub-Task 1.1, we can observe that teams with negative GAP values considered Diffusion Models for the generation of synthetic data (\textit{i.e.,} ID R\&D uses HDT, SRCN\_AIVL combines DCFace and IDiff-Face, and CTAI combines DCFace and GANDiffFace), showing that this generation method may work better than real data in scenarios with limited data. Next, after removing the limitation on the number of synthetic images (\textit{i.e.,} Sub-Task 1.2), the TO value of most FR systems increases, which leads to performance and fairness improvement simultaneously. For instance, for the ADMIS team (ranked top-2 in both Sub-Task 1.1 and 1.2), the TO value increases to 95.72\% in Sub-Task 1.2 (\textit{i.e.,} 1.42\% TO improvement compared to Sub-Task 1.1). Also, the GAP value decreases from 1.47\% to -0.56\%, obtaining better results when increasing the amount of synthetic data in comparison to limited real data (i.e., CASIA-WebFace). Another example is the OPDAI team, which raised from top-4 to top-3 positions between Sub-Task 1.1 and 1.2. Its TO value increases to 94.12\% (\textit{i.e.,} 0.37\% TO improvement from Sub-Tasks 1.1 to 1.2), and the GAP value is reduced from 1.02\% to 0.71\%. These findings emphasize the potential of generating large number of synthetic face images from different demographic groups to mitigate bias in existing FR technology. Finally, we analyze in Sub-Task 1.3 the case of using both, real and synthetic data, in the FR training process. In general, we can observe considerable improvements in terms of TO values, along with higher negative GAP values for all the top-6 teams, \textit{e.g.,} ADMIS (96.50\% TO, -1.33 GAP), K-IBS-DS (96.17\% TO, -1.37\% GAP), and UNICA-IGD-LSI (96.00\% TO, -5.33\% GAP). Moreover, Table~\ref{tab:task1_baselinevsproposed} shows the performance of each demographic group for the baseline models, \textit{i.e.,} those trained only with real data, and the models proposed by the top-ranked teams for each of the sub-tasks of Task 1. The ID R\&D team achieves higher performance across all demographic groups except for white males and females (98.40\% vs. 97.90\% AVG, and 97.60\% vs. 96.50\% AVG, respectively). We believe this is produced due to the class imbalance in the real data for this demographic group, which also leads to achieving the best performance for the case of training only with real data, due to overfitting. Similarly, the ADMIS team achieves higher performance across all demographic groups except for black males (98.00\% vs. 97.10\% AVG). Moreover, when we analyze the scenario of training only with synthetic data, not only does the overall performance increase, but the bias between demographic groups is also reduced (0.82\% vs 4.40\% STD for Sub-Tasks 1.1 and 1.2, and 0.75\% vs 1.81\% STD for Sub-Task 1.3), which implies a better model in terms of fairness across all demographic groups. This is mainly produced as with synthetic data we can control how balanced the data is with respect to these groups. Finally, it is noteworthy to compare the best results achieved in Sub-Task 1.2, \textit{i.e.,} unconstrained synthetic data, and Sub-Task 1.3, \textit{i.e.,} constrained synthetic + real data. The ID R\&D team achieves 96.73\% TO in Sub-Task 1.2, whereas ADMIS achieves 96.50\% TO in Sub-Task 1.3, showing that unlimited synthetic data for training can even outperform FR systems trained with limited synthetic + real data. These results motivate the use of synthetic data for demographic bias mitigation, improving at the same time privacy as no real identities are seen by the network.

\begin{table*}[t!]
\caption{Ranking for the three sub-tasks considered in Task 2. For each sub-task, we highlight in bold the best team according to the Average Accuracy. AVG = Average accuracy, FNMR = False Non-Match Rate, FMR = False Match Rate, AUC = Area Under Curve, GAP = Gap to Real.}
\label{tab:task2}
\centering
\resizebox{0.8\linewidth}{!}{\begin{tabular}{@{}llcccc@{}}
\toprule
\multicolumn{6}{c}{\textbf{Sub-Task   2.1 (Overall Improvement): Synthetic Data (Constrained)}} \\ \midrule
Pos. & Team & AVG {[}\%{]} & FNMR@FMR=1\% & AUC {[}\%{]} & GAP {[}\%{]} \\ \midrule
\textbf{1} & \textbf{OPDAI} & \textbf{91.93} & \textbf{17.63 (2)} & \textbf{97.30 (2)} & \textbf{3.09} \\
2 & ID R\&D & 91.86 & 10.36 (1) & 97.48 (1) & 2.99 \\
3 & ADMIS & 91.19 & 20.41 (3) & 97.04 (3) & 2.78 \\
4 & K-IBS-DS & 91.05 & 24.87 (6) & 96.09 (6) & 2.60 \\
5 & CTAI & 90.59 & 21.88 (4) & 96.40 (5) & -1.94 \\
6 & BOVIFOCR-UFPR & 89.97 & 24.04 (5) & 96.70 (4) & 3.71 \\ \bottomrule
\end{tabular}}

\bigskip

\resizebox{0.8\linewidth}{!}{\begin{tabular}{@{}llcccc@{}}
\toprule
\multicolumn{6}{c}{\textbf{Sub-Task 2.2 (Overall Improvement): Synthetic Data   (Unconstrained)}} \\ \midrule
Pos. & Team & AVG {[}\%{]} & FNMR@FMR=1\% & AUC {[}\%{]} & GAP {[}\%{]} \\ \midrule
\textbf{1} & \textbf{Idiap-SynthDistill} & \textbf{93.50} & \textbf{9.17 (1)} & \textbf{97.17 (4)} & \textbf{-0.05} \\
2 & ADMIS & 92.92 & 14.45 (3) & 97.76 (1) & 0.21 \\
3 & OPDAI & 92.04 & 16.48 (4)& 97.41 (3) & 3.00 \\
4 & ID R\&D & 91.86 & 10.36 (2)& 97.48 (2) & 2.99 \\
5 & K-IBS-DS & 91.61 & 22.48 (6)& 96.54 (5) & 1.96 \\
6 & CTAI & 90.59 & 21.88 (5)& 96.40 (6) & -1.94 \\ \bottomrule
\end{tabular}}

\bigskip

\resizebox{0.8\linewidth}{!}{\begin{tabular}{@{}llcccc@{}}
\toprule
\multicolumn{6}{c}{\textbf{Sub-Task  2.3 (Overall Improvement): Synthetic + Real Data (Constrained)}} \\ \midrule
Pos. & Team & AVG {[}\%{]} & FNMR@FMR=1\% & AUC {[}\%{]} & GAP {[}\%{]} \\ \midrule
\textbf{1} & \textbf{K-IBS-DS} & \textbf{95.42} & \textbf{9.49 (5)} & \textbf{98.14 (6)} & \textbf{-2.15} \\
2 & OPDAI & 95.23 & 7.54 (1)& 98.70 (1) & -0.52 \\
3 & CTAI & 94.56 & 8.85 (4) & 98.41 (3) & -6.01 \\
4 & CBSR-Samsung & 94.20 & 8.62 (3)& 98.17 (4) & -4.40 \\
5 & ADMIS & 94.15 & 10.99 (6)& 98.46 (2) & -1.10 \\
6 & ID R\&D & 94.05 & 8.00 (2)& 98.16 (5) & 0.07 \\ \bottomrule
\end{tabular}
}
\end{table*}

\begin{table*}[t]
\caption{Comparison between the baseline model (trained exclusively with real data), and the proposed model (using synthetic data) for each finalist of Task 2. The performance of each database is represented by its AVG {[}\%{]}.}
\label{tab:task2_baselinevsproposed}
\centering
\resizebox{\linewidth}{!}{\begin{tabular}{cccccccccc}
\hline
\textbf{Sub-Task} & \textbf{Team} & \textbf{Model} & \textbf{AgeDB} & \textbf{BUPT} & \textbf{CFP-FP} & \textbf{ROF} & \textbf{AVG {[}\%{]}} & \textbf{STD {[}\%{]}} & \textbf{GAP {[}\%{]}} \\ \hline
\multirow{2}{*}{2.1} & \multirow{2}{*}{OPDAI} & Proposed & 94.54 & 93.42 & 92.51 & 87.22 & 91.93 & 2.81 & \multirow{2}{*}{3.09} \\ \cline{3-9}
 &  & Baseline & \textbf{96.78} & \textbf{93.47} & \textbf{97.29} & \textbf{91.39} & \textbf{94.73} & \textbf{2.42} &  \\ \hline
\multirow{2}{*}{2.2} & \multirow{2}{*}{\begin{tabular}[c]{@{}c@{}}Idiap\\ SynthDistill\end{tabular}} & Proposed & 96.72 & 93.85 & \textbf{96.14} & \textbf{87.31} & 93.5 & \textbf{3.74} & \multirow{2}{*}{-0.05} \\ \cline{3-9}
 &  & Baseline & \textbf{97.6} & \textbf{97.22} & 95.1 & 84.14 & \textbf{93.52} & 5.5 &  \\ \hline
\multirow{2}{*}{2.3} & \multirow{2}{*}{K-IBS-DS} & Proposed & \textbf{96.89} & \textbf{96.88} & \textbf{97.51} & \textbf{90.39} & \textbf{95.42} & \textbf{2.91} & \multirow{2}{*}{-2.16} \\ \cline{3-9}
 &  & Baseline & 96.2 & 93.76 & 95.97 & 87.58 & 93.38 & 3.48 &  \\ \hline
\end{tabular}
}
\end{table*}

\subsection{Task 2: Overall Improvement}\label{subsec:R_T2}
Table~\ref{tab:task2} provides the results achieved by participants in Task 2, focusing not only on bias mitigation but also other challenges in FR such as age, pose, and occlusions. Teams are ranked in descending order based on the average verification accuracy across the four databases. Notably, in all sub-tasks, the AVG is lower than the achieved in Task 1 for the BUPT-BalancedFace~\cite{wang2020mitigating} database, showing the additional challenges introduced by the AgeDB~\cite{moschoglou2017agedb}, CFP-FP~\cite{sengupta2016frontal}, and ROF~\cite{erakιn2021recognizing} databases. This trend can also be observed in GAP results, which tend to be worse for Sub-Tasks 2.1 and 2.2 compared to Sub-Tasks 1.1 and 1.2, suggesting that it is far more difficult to emulate the conditions of these real databases with synthetic data by itself. For example, in Sub-Task 2.1, although the top-3 teams achieve high AVG results, they exhibit a considerable positive GAP value (\textit{i.e.,} OPDAI 91.93\% AVG, 3.09\% GAP; ID R\&D 91.86\% AVG, 2.99\% GAP; ADMIS 91.19\% AVG, 2.78\% GAP), showing that a FR model trained only with real data (\textit{i.e.,} CASIA-WebFace) adapts better to adverse image conditions such as aging, pose, or occlusions. Focusing on Sub-Task 2.2, the team ranked top-1, \textit{i.e.,} Idiap-SynthDistill, achieves much better results compared to the best result of Sub-Task 2.1 (\textit{i.e.,} 93.50\% vs. 91.93\% AVG), proving that unlimited synthetic data can further improve the performance of the system. Finally, in Sub-Task 2.3, most teams report better AVG and higher negative GAP values (\textit{e.g.,} K-IBS-DS achieves 95.42\% AVG, -2.15\% GAP), proving again that synthetic data combined with real data can alleviate existing limitations within FR technology.

Furthermore, Table~\ref{tab:task2_baselinevsproposed} compares the performance across the different evaluation databases, highlighting the variations between the baseline models, which were trained only on real data, and the models proposed by the top-ranked teams for the sub-tasks of Task 2.In Sub-Task 2.1, where the model proposed by OPDAI is trained with limited synthetic data due to competition rules, it underperforms compared to the baseline across all databases (3.09\% GAP). This suggests that 500,000 synthetic samples are insufficient to fully replace real data. Moreover, in Sub-Task 2.2, where training synthetic data is unlimited, the performance of the model proposed by Idiap-SynthDistill is almost identical to the baseline (-0.05\% GAP). This suggests that a large synthetic dataset can closely replicate the distribution of real data. We can also observe that in AgeDB and BUPT databases, the baseline still outperforms the proposed model, which indicates that synthetic data does not fully capture the differences between aging or demographic groups. Finally, in Sub-Task 2.3, which combines real and unlimited synthetic data, the model proposed by K-IBS-DS outperforms the baseline in all databases (-2.16\% GAP). This supports the idea of all our previous experiments, that synthetic data complements real data, improving generalization.

\subsection{Demographic Groups and Evaluation Databases}\label{subsec:R_Dem}
\begin{figure*}[t!]
\centering
\begin{tabular}{cc}
    \includegraphics[width=0.49\linewidth]{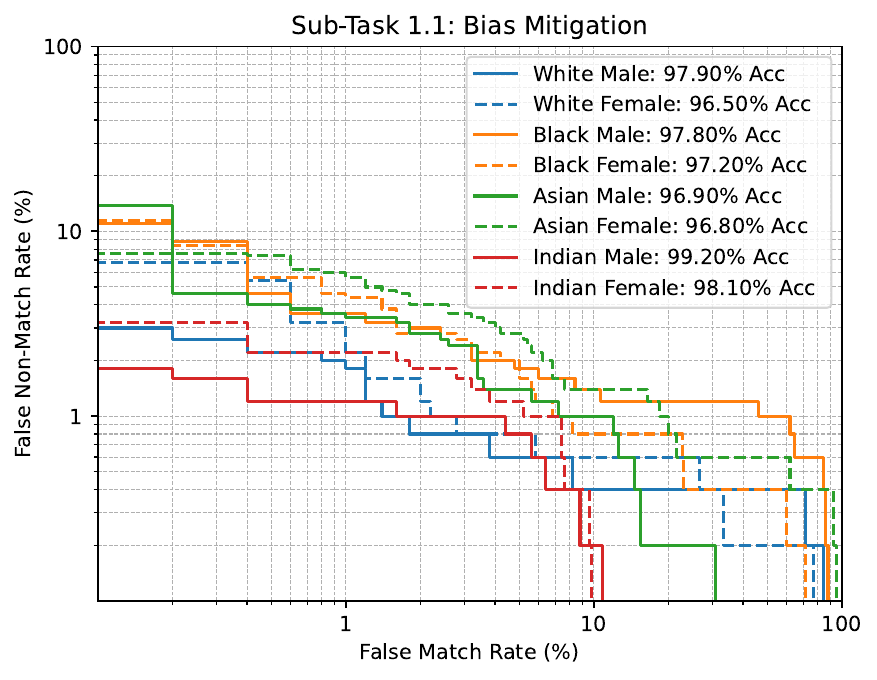} &   \includegraphics[width=0.49\linewidth]{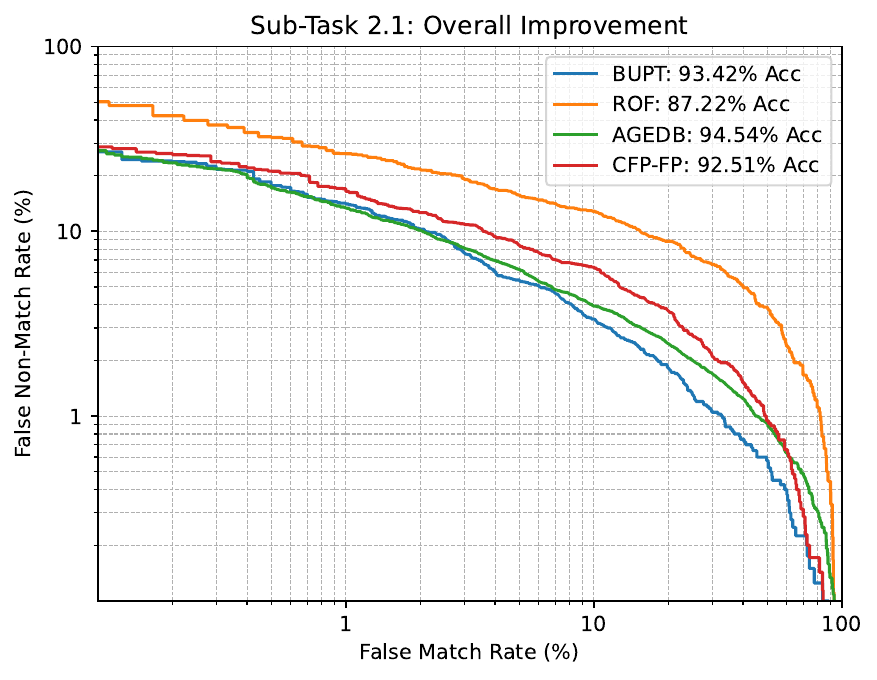}\\
    \includegraphics[width=0.49\linewidth]{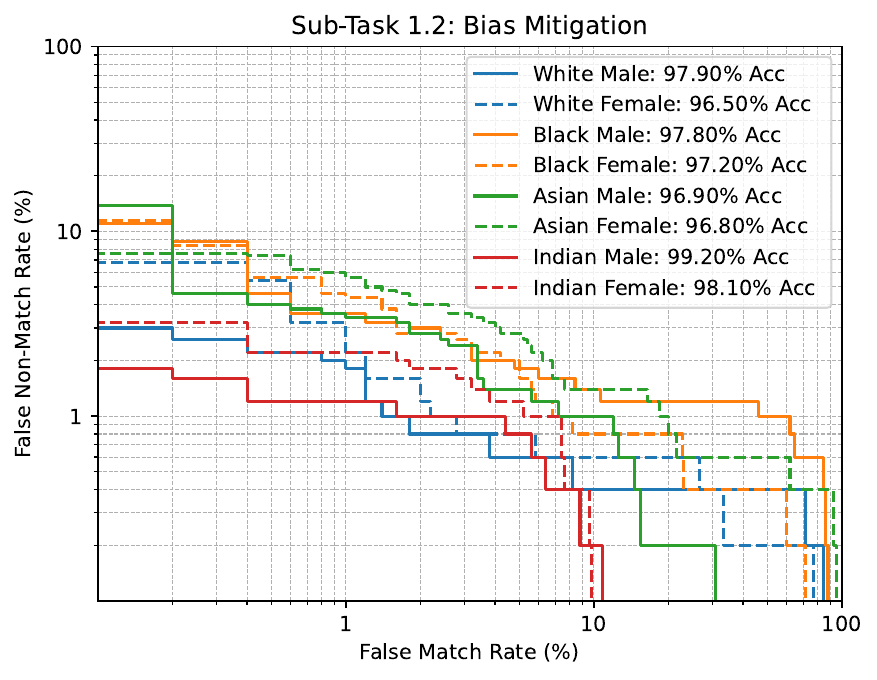} & \includegraphics[width=0.49\linewidth]{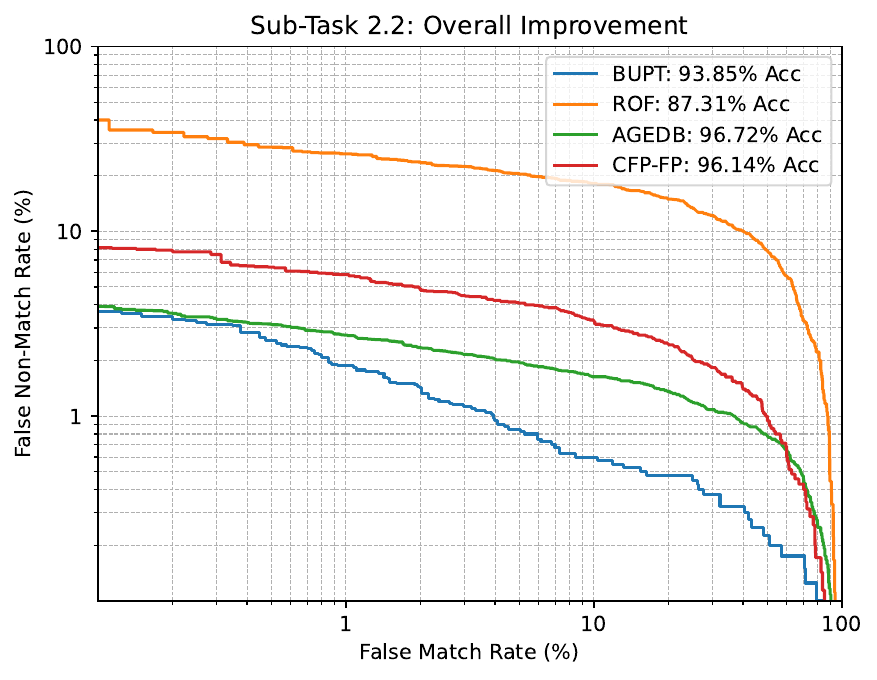} \\
    \includegraphics[width=0.49\linewidth]{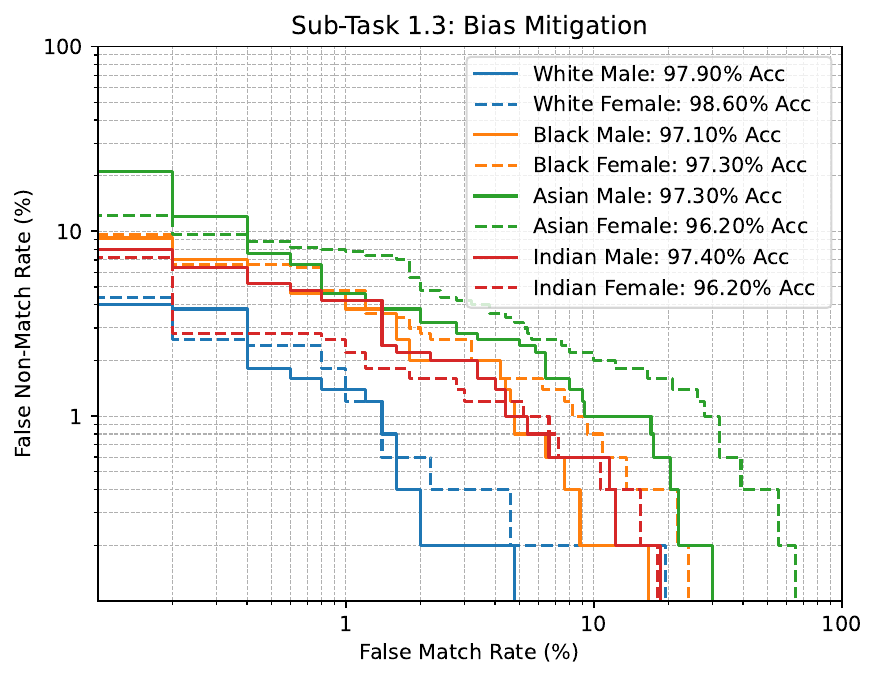} & \includegraphics[width=0.49\linewidth]{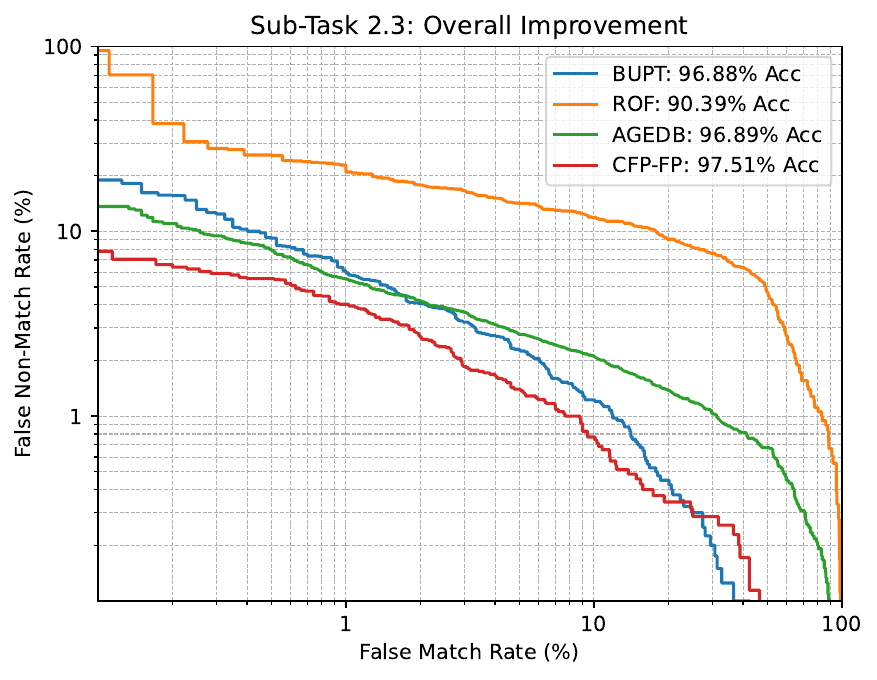} \\
    
\end{tabular}
\caption{DET curves of Task 1 (left) and Task 2 (right). Sub-Tasks 1.1 (left-top), 1.2 (left-middle), and 1.3 (left-bottom), including the results achieved for the top-1 team (i.e., ID R\&D, ID R\&D, ADMIS, respectively) in each demographic group. Sub-Tasks 2.1 (right-top), 2.2 (right-middle), and 2.3 (right-bottom), including the results achieved for the top-1 team (i.e., OPDAI, Idiap-SynthDistill, K-IBS-DS, respectively) in each demographic group.}
\label{fig:DET}
\end{figure*}

This section provides an in-depth analysis of the results in terms of the different demographic groups and individual databases considered in the 2$^\text{nd}$ FRCSyn-onGoing. Figure~\ref{fig:DET} (left), shows the Detection Error Tradeoff (DET) curves of Sub-Tasks 1.1, 1.2, and 1.3, including the results achieved for the top-1 team in each demographic group. For completeness, the information and graphical representations for all the teams can be found on the challenge Codalab platform\footnote{\href{https://codalab.lisn.upsaclay.fr/competitions/16970\#results}{https://codalab.lisn.upsaclay.fr/competitions/16970\#results}}.

For Sub-Tasks 1.1 and 1.2, the team that achieves the first place, ID R\&D, demonstrates high performance across the different demographic groups considered (above 96.50\% Accuracy for all demographic groups). However, a slight gender bias can be observed (improvements of $\approx$1\% between `Male' and `Female' labels for some ethnicities). Regarding the ethnicity, the proposed FR model showed better results for subjects from the `Indian' ethnicity (99.20\% Accuracy for Indian Male; 98.10\% Accuracy for Indian Female). Finally, for Sub-Task 1.3, the winning team, ADMIS, also performs well across all demographic groups (all above 96\% Accuracy). However, there exists variability in performance between different demographic groups. For example, `Asian Females' and `Indian Females' have the lowest Accuracy (96.20\%) while `White Females' have the highest Accuracy (98.60\%).

Figure~\ref{fig:DET} (right) shows the DET curves of Sub-Tasks 2.1, 2.2, and 2.3, including the results achieved for the top-1 team in each database. Analyzing the FR model proposed by the OPDAI team for Sub-Task 2.1, the spread of the curves indicates variability in the system performance across different databases, with the results from AgeDB (94.54\% Accuracy) outperforming others. Moreover, in Sub-Task 2.2 the Idiap-SynthDistill FR model significantly improves the performance of Sub-Task 2.1 for AgeDB and CFP-FP databases (\textit{i.e.,} 96.72\% and 96.14\% Accuracy, respectively). Finally, for Sub-Task 2.3, the curves from the K-IBS-DS FR model are closely aligned for the AgeDB (96.89\% Accuracy), BUPT (96.88\% Accuracy), and CFP-FP (97.51\% Accuracy), showing consistent and reliable performance across these databases. However, the curve from the ROF database remains the worst in each sub-task (\textit{i.e.,} 87.22\%, 87.31\%, and 90.39\% Accuracy for Sub-Tasks 2.1, 2.2, and 2.3, respectively), reflecting that it is the most difficult database to emulate with synthetic data.

\subsection{Post-Challenge Analysis and Comparison with 1$^\text{st}$ Edition}\label{subsec:RCMP}

Analyzing the contributions of all eleven top teams, we can observe the prevalence of well-established methodologies. Notably, most teams used DCFace~\cite{kim2023dcface} either independently or in conjunction with other synthetic databases such as GANDiffFace~\cite{melzi2023gandiffface}, DigiFace-1M~\cite{bae2023digiface}, or IDiff-Face~\cite{boutros2023idiff}. DCFace is a Dual Condition Face Generator based on a diffusion model, designed to create facial images of the same subject in various styles while maintaining identity consistency. Its key component is the \textit{Patch-wise Style Extractor}, which extracts style features from an image while minimizing identity information. This forces the model to rely on a separate input for identity data. Unlike previous approaches like StyleGAN~\cite{karras2019style}, DCFace retains essential spatial details, such as pose, ensuring greater variability between subjects of the same identity. This results in images with a similar style that enhance the performance of facial recognition models in identifying subjects. Furthermore, several teams, including CBSR-Samsung, INESC-IGD, and CTAI adopted interesting approaches involving synthetic data cleaning and selection. These approaches include: \textit{i)} de-overlapping the data from DCFace, as it is trained with CASIA-WebFace and some data could be very similar, deteriorating the training, \textit{ii)} balancing the data with respect to the demographic information, and \textit{iii)} removing the images that are far from the class center using clustering techniques such as DBSCAN. It is interesting to highlight the ID R\&D and Idiap-SynthDistill teams as they considered novel methods to generate synthetic data. Specifically, the ID R\&D team used an HDT~\cite{crowson2024scalable} to generate synthetic facial images along with identity and style embeddings, which were used by a StyleNAT~\cite{walton2022stylenat} model to generate more variability in the synthetic data. Another example is the Idiap-SynthDistill team, which proposed an end-to-end method that dynamically generated facial images through StyleGAN2~\cite{otroshishahreza2024knowledge} and trained a FR model through model distillation. Regarding the backbone architecture, all teams opted for either ResNet~\cite{he2016deep} or IResNet~\cite{duta2021improved}, mainly for their widespread adoption in state-of-the-art FR methodologies. Both architectures use residual connections to improve the training, but while in ResNet, the skip connections bypass one or more layers to address the vanishing gradient problem, IResNet optimizes the information flow through the network, allowing for the training of extremely deep architectures without increasing model complexity. Finally, the selection of the loss functions was also similar among the teams, with AdaFace~\cite{kim2022adaface} and ArcFace~\cite{deng2019arcface} the prevalent choices. Both losses use angular margin-based loss functions to improve facial feature discrimination, but while in ArcFace the margin is fixed for all samples, AdaFace adapts the margin dynamically based on the quality of each image. Nevertheless, there were exceptions such as for the ID R\&D team that used the recent UniFace~\cite{zhou2023uniface}, or the UNICA team that considered CosFace~\cite{wang2018cosface}.

\begin{table*}[t]
\caption{Description of the best results achieved in the 1$^\text{st}$ and 2$^\text{nd}$ FRCSyn-onGoing. Sub-Tasks 1.2 and 2.2 of the 2$^\text{nd}$ FRCSyn-onGoing are not included in the table as they are novel sub-tasks only available in the 2$^\text{nd}$ edition.}
\label{tab:comparison}
\begin{tabular}{cc}
\begin{minipage}{0.45\textwidth}
\centering
\resizebox{\textwidth}{!}{\begin{tabular}{@{}cccccc@{}}
\toprule
\multicolumn{6}{c}{\textbf{Sub-Task 1.1: Bias Mitigation}} \\ \midrule
\multicolumn{3}{c}{\textbf{1$^\text{st}$ FRCSyn-onGoing}} & \multicolumn{3}{c}{\textbf{2$^\text{nd}$ FRCSyn-onGoing}} \\ \midrule
Team & TO {[}\%{]} & GAP {[}\%{]} & Team & TO {[}\%{]} & GAP {[}\%{]} \\ \midrule
LENS & 92.25 & -0.74 & ID R\&D & \textbf{96.73} & \textbf{-5.31} \\ \bottomrule
\end{tabular}}
\bigskip
\resizebox{\textwidth}{!}{\begin{tabular}{@{}cccccc@{}}
\toprule
\multicolumn{6}{c}{\textbf{Sub-Task 1.3: Bias Mitigation}} \\ \midrule
\multicolumn{3}{c}{\textbf{1$^\text{st}$ FRCSyn-onGoing}} & \multicolumn{3}{c}{\textbf{2$^\text{nd}$ FRCSyn-onGoing}} \\ \midrule
Team & TO {[}\%{]} & GAP {[}\%{]} & Team & TO {[}\%{]} & GAP {[}\%{]} \\ \midrule
CBSR & 95.25 & \textbf{-2.10} & ADMIS & \textbf{96.50} & -1.33 \\ \bottomrule
\end{tabular}}
\end{minipage}

&

\begin{minipage}{0.52\textwidth}
\centering
\resizebox{\textwidth}{!}{\begin{tabular}{@{}cccccc@{}}
\toprule
\multicolumn{6}{c}{\textbf{Sub-Task 2.1: Overall Improvement}} \\ \midrule
\multicolumn{3}{c}{\textbf{1$^\text{st}$ FRCSyn-onGoing}} & \multicolumn{3}{c}{\textbf{2$^\text{nd}$ FRCSyn-onGoing}} \\ \midrule
Team & AVG {[}\%{]} & GAP {[}\%{]} & Team & AVG {[}\%{]} & GAP {[}\%{]} \\ \midrule
BOVIFOCR & 90.50 & \textbf{2.66} & OPDAI & \textbf{91.93} & 3.09 \\ \bottomrule
\end{tabular}}
\bigskip
\resizebox{\textwidth}{!}{\begin{tabular}{@{}cccccc@{}}
\toprule
\multicolumn{6}{c}{\textbf{Sub-Task 2.3: Overall Improvement}} \\ \midrule
\multicolumn{3}{c}{\textbf{1$^\text{st}$ FRCSyn Challenge}} & \multicolumn{3}{c}{\textbf{2$^\text{nd}$ FRCSyn-onGoing}} \\ \midrule
Team & AVG {[}\%{]} & GAP {[}\%{]} & Team & AVG {[}\%{]} & GAP {[}\%{]} \\ \midrule
BOVIFOCR & 94.95 & \textbf{-3.69} & K-IBS-DS & \textbf{95.42} & -2.15 \\ \bottomrule
\end{tabular}}
\end{minipage}
\end{tabular}
\end{table*}

Aditionally, we compare the results achieved in the 2$^\text{nd}$ FRCSyn-onGoing with the results of the 1$^\text{st}$ edition~\cite{melzi2024frcsyn}. Table~\ref{tab:comparison} shows the best results achieved in the 1$^\text{st}$ and 2$^\text{nd}$ editions of the challenge, including also the GAP values. It is important to remark that Sub-Tasks 1.2 and 2.2 of the 2$^\text{nd}$ FRCSyn-onGoing are not included in the analysis as they are novel sub-tasks only available in the 2$^\text{nd}$ edition. Notably, two observations can be made: \textit{i)} the main metric for ranking teams (\textit{i.e.,} TO and AVG) shows improvements across all cases in this 2$^\text{nd}$ edition for both Task 1 (\textit{e.g.,} 96.73\% vs. 92.25\% TO in Sub-Task 1.1 and 96.50\% vs. 95.25\% TO in Sub-Task 1.3) and Task 2 (\textit{e.g.,} 91.93\% vs. 90.50\% AVG in Sub-Task 2.1 and 95.42\% vs. 94.95\% AVG in Sub-Task 2.3), and \textit{ii)} in terms of the GAP value, the FR models of this 2$^\text{nd}$ edition follow a similar trend compared to the 1$\text{st}$ edition, achieving in most sub-tasks negative GAP values, remarking the benefits for training using synthetic data. In particular, for Sub-Task 1.1, a much higher negative GAP value is observed in this 2$^\text{nd}$ edition (\textit{i.e.,} -5.31\% vs. -0.74\%). This result, together with the higher TO value, seems to be due to the generation of better synthetic data by the ID R\&D team, with the proposal of novel generative methods, as indicated before. In addition, several conclusions can be drawn. First, the improvement in the main metric can be associated with the freedom to select the methodology to generate the synthetic data to train the FR models, as well as the application of data cleaning and selection techniques. We observe that the increasing GAP value can be associated also with the enhancement of the FR models, due to the proposal of different architectures and loss functions, as indicated before.

Finally, we observe that when it comes to the quality of the generated synthetic data, higher quality does not imply better performance in recognition tasks. Most teams use DCFace as the main dataset for their training, even though it generates images with lower resolution and less detail from a qualitative perspective. Nevertheless, analyzing the results achieved using this database, we conclude that FR models might not require highly detailed images to learn to match identities, at least for the FR databases considered in this challenge. This suggests that instead of focusing on realism, synthetic data for FR should generate diverse images that help models better learn class variability while reducing noise.

\section{Conclusion} \label{sec:conclusion}
The 2$^\text{nd}$ FRCSyn-onGoing has presented a comprehensive exploration of the applications of synthetic data in FR, effectively addressing existing limitations in the field. In this 2$^\text{nd}$ edition, two additional sub-tasks have been introduced, showing that impressive results can be achieved using unlimited synthetic data, even outperforming in some cases the scenario of training with only real data. With an increased number of participants in this last edition, we have witnessed a considerable performance improvement in all sub-tasks in comparison to the 1$^\text{st}$ edition~\cite{melzi2024frcsyn, melzi2024frcsyna}. This has been possible thanks to the proposal of novel methods to generate and select better synthetic data, as well as FR models and loss functions. These approaches can be compared across a variety of sub-tasks, with many being reproducible thanks to the materials made available by the participating teams. 

Future studies will include recent AI techniques~\cite{dealcala2024is}, to make sure that only the databases available by the challenge are used by participants. We will also perform a more detailed analysis of the results and comparison with recent challenges in the topic, such as SDFR~\cite{shahreza2024sdfr}, or evaluate over a more diverse set of databases that include other FR challenges, like quality, surveillance, or large distance images~\cite{liu2024farsight}. Finally, we plan to focus on the explainability of these FR models and the frameworks that generate synthetic images~\cite{deandrestame2024how}. Debiasing face recognition models using synthetic datasets is an important task and in this challenge, we found that the use of synthetic data can furhter increase the performance. However, the concept of ``bias" itself is complex and often subjective. What constitutes a ``fair" representation can vary significantly depending on cultural context, individual experiences, and even personal beliefs. Therefore, debiasing efforts should be approached with an accurate understanding of the multifaceted nature of bias. Simply generating synthetic data to reflect a particular demographic distribution might not fully address the complexities of real-world inequalities. Rather than seeking to eliminate bias, perhaps a more productive approach is to pursue research in the direction of transparency, interpretability, and controllability. This translates to research questions that allow researchers to easily define what bias should be and allow them to fine-tune their models accordingly. Ultimately, the goal should be to develop face recognition systems that are not only accurate but also fair and ethical.

\section*{Acknowledgements}
{\footnotesize This study is supported by INTER-ACTION (PID2021-126521OB-I00 MICINN/FEDER), Cátedra ENIA UAM-VERIDAS en IA Responsable (NextGenerationEU PRTR TSI-100927-2023-2), R\&D Agreement DGGC/UAM/FUAM for Biometrics and Cybersecurity, and PowerAI+ (SI4/PJI/2024-00062, funded by Comunidad de Madrid through the grant agreement for the promotion of research and technology transfer at UAM). 
It is also supported by the German Federal Ministry of Education and Research and the Hessian Ministry of Higher Education, Research, Science, and the Arts within their joint support of the National Research Center for Applied Cybersecurity ATHENE.
K-IBS-DS was supported by the Institute for Basic Science, Republic of Korea (IBS-R029-C2). UNICA-IGD-LSI was supported by the ARIS program P2-0250B.}

\bibliographystyle{IEEEtran}
\bibliography{egbib}

\begin{thebibliography}{10}
\providecommand{\url}[1]{#1}
\csname url@samestyle\endcsname
\providecommand{\newblock}{\relax}
\providecommand{\bibinfo}[2]{#2}
\providecommand{\BIBentrySTDinterwordspacing}{\spaceskip=0pt\relax}
\providecommand{\BIBentryALTinterwordstretchfactor}{4}
\providecommand{\BIBentryALTinterwordspacing}{\spaceskip=\fontdimen2\font plus
\BIBentryALTinterwordstretchfactor\fontdimen3\font minus \fontdimen4\font\relax}
\providecommand{\BIBforeignlanguage}[2]{{%
\expandafter\ifx\csname l@#1\endcsname\relax
\typeout{** WARNING: IEEEtran.bst: No hyphenation pattern has been}%
\typeout{** loaded for the language `#1'. Using the pattern for}%
\typeout{** the default language instead.}%
\else
\language=\csname l@#1\endcsname
\fi
#2}}
\providecommand{\BIBdecl}{\relax}
\BIBdecl

\bibitem{wang2021deep}
M.~Wang and W.~Deng, ``{Deep Face Recognition: A Survey},'' \emph{Neurocomputing}, vol. 429, pp. 215--244, 2021.

\bibitem{du2022elements}
H.~Du, H.~Shi, D.~Zeng, X.-P. Zhang, and T.~Mei, ``{The Elements of End-to-end Deep Face Recognition: A Survey of Recent Advances},'' \emph{ACM Comput. Surv.}, vol.~54, no. 10s, pp. 1--42, 2022.

\bibitem{bisogni2022impact}
C.~Bisogni, A.~Castiglione, S.~Hossain, F.~Narducci, and S.~Umer, ``{Impact of Deep Learning Approaches on Facial Expression Recognition in Healthcare Industries},'' \emph{IEEE Transactions on Industrial Informatics}, vol.~18, no.~8, pp. 5619--5627, 2022.

\bibitem{gomez2021improving}
L.~F. Gomez, A.~Morales, J.~R. Orozco-Arroyave, R.~Daza, and J.~Fierrez, ``{Improving Parkinson Detection Using Dynamic Features From Evoked Expressions in Video},'' in \emph{Proc. IEEE/CVF Conference on Computer Vision and Pattern Recognition Workshops}, 2021.

\bibitem{daza2023matt}
R.~Daza, L.~F. Gomez, A.~Morales, J.~Fierrez, R.~Tolosana, R.~Cobos, and J.~Ortega-Garcia, ``{MATT: Multimodal Attention Level Estimation for e-learning Platforms},'' in \emph{Proc. AAAI Workshop on Artificial Intelligence for Education}, 2023.

\bibitem{deng2019arcface}
J.~Deng, J.~Guo, N.~Xue, and S.~Zafeiriou, ``{ArcFace: Additive Angular Margin Loss for Deep Face Recognition},'' in \emph{Proc. IEEE/CVF Conference on Computer Vision and Pattern Recognition}, 2019.

\bibitem{kim2022adaface}
M.~Kim, A.~K. Jain, and X.~Liu, ``{AdaFace: Quality Adaptive Margin for Face Recognition},'' in \emph{Proc. IEEE/CVF Conference on Computer Vision and Pattern Recognition}, 2022.

\bibitem{deandrestame2024how}
I.~Deandres-Tame, R.~Tolosana, R.~Vera-Rodriguez, A.~Morales, J.~Fierrez, and J.~Ortega-Garcia, ``{How Good Is ChatGPT at Face Biometrics? A First Look Into Recognition, Soft Biometrics, and Explainability},'' \emph{IEEE Access}, vol.~12, pp. 34\,390--34\,401, 2024.

\bibitem{crum2023explain}
C.~R. Crum, P.~Tinsley, A.~Boyd, J.~Piland, C.~Sweet, T.~Kelley, K.~Bowyer, and A.~Czajka, ``{Explain To Me: Salience-Based Explainability for Synthetic Face Detection Models},'' \emph{IEEE Transactions on Artificial Intelligence}, pp. 1--12, 2023.

\bibitem{shen2022interfacegan}
Y.~Shen, C.~Yang, X.~Tang, and B.~Zhou, ``{InterFaceGAN: Interpreting the Disentangled Face Representation Learned by GANs},'' \emph{IEEE Transactions on Pattern Analysis and Machine Intelligence}, vol.~44, no.~4, pp. 2004--2018, 2022.

\bibitem{terhoerst2021comprehensive}
P.~Terh{\"o}rst, J.~N. Kolf, M.~Huber, F.~Kirchbuchner, N.~Damer, A.~M. Moreno, J.~Fierrez, and A.~Kuijper, ``{A Comprehensive Study on Face Recognition Biases Beyond Demographics},'' \emph{IEEE Transactions on Technology and Society}, vol.~3, no.~1, pp. 16--30, 2021.

\bibitem{melzi2023synthetic}
P.~Melzi, C.~Rathgeb, R.~Tolosana, R.~Vera-Rodriguez, A.~Morales, D.~Lawatsch, F.~Domin, and M.~Schaubert, ``{Synthetic Data for the Mitigation of Demographic Biases in Face Recognition},'' in \emph{Proc. IEEE International Joint Conference on Biometrics}, 2023.

\bibitem{morales2021sensitivenets}
A.~Morales, J.~Fierrez, R.~Vera-Rodriguez, and R.~Tolosana, ``{SensitiveNets: Learning Agnostic Representations with Application to Face Images},'' \emph{IEEE Transactions on Pattern Analysis and Machine Intelligence}, vol.~43, no.~6, pp. 2158--2164, 2021.

\bibitem{melzi2023multi}
P.~Melzi, H.~O. Shahreza, C.~Rathgeb, R.~Tolosana, R.~Vera-Rodriguez, J.~Fierrez, S.~Marcel, and C.~Busch, ``{Multi-IVE: Privacy Enhancement of Multiple Soft-Biometrics in Face Embeddings},'' in \emph{Proc. IEEE/CVF Winter Conference on Applications of Computer Vision Workshops}, 2023.

\bibitem{melzi2024overview}
P.~Melzi, C.~Rathgeb, R.~Tolosana, R.~Vera, and C.~Busch, ``{An Overview of Privacy-Enhancing Technologies in Biometric Recognition},'' \emph{ACM Computing Surveys}, 2024.

\bibitem{zhao2022towards}
J.~Zhao, S.~Yan, and J.~Feng, ``{Towards Age-Invariant Face Recognition},'' \emph{IEEE Transactions on Pattern Analysis and Machine Intelligence}, vol.~44, no.~1, pp. 474--487, 2022.

\bibitem{valle2021multi}
R.~Valle, J.~M. Buenaposada, and L.~Baumela, ``{Multi-Task Head Pose Estimation in-the-Wild},'' \emph{IEEE Transactions on Pattern Analysis and Machine Intelligence}, vol.~43, no.~8, pp. 2874--2881, 2021.

\bibitem{tran2019representation}
L.~Tran, X.~Yin, and X.~Liu, ``{Representation Learning by Rotating Your Faces},'' \emph{IEEE Transactions on Pattern Analysis and Machine Intelligence}, vol.~41, no.~12, pp. 3007--3021, 2019.

\bibitem{mudunuri2016low}
S.~P. Mudunuri and S.~Biswas, ``{Low Resolution Face Recognition Across Variations in Pose and Illumination},'' \emph{IEEE Transactions on Pattern Analysis and Machine Intelligence}, vol.~38, no.~5, pp. 1034--1040, 2016.

\bibitem{qiu2022end2end}
H.~Qiu, D.~Gong, Z.~Li, W.~Liu, and D.~Tao, ``{End2End Occluded Face Recognition by Masking Corrupted Features},'' \emph{IEEE Transactions on Pattern Analysis and Machine Intelligence}, vol.~44, no.~10, pp. 6939--6952, 2022.

\bibitem{melzi2023gandiffface}
P.~Melzi, C.~Rathgeb, R.~Tolosana, R.~Vera-Rodriguez, D.~Lawatsch, F.~Domin, and M.~Schaubert, ``{GANDiffFace: Controllable Generation of Synthetic Datasets for Face Recognition with Realistic Variations},'' in \emph{Proc. IEEE/CVF International Conference on Computer Vision Workshops}, 2023.

\bibitem{kim2023dcface}
M.~Kim, F.~Liu, A.~Jain, and X.~Liu, ``{DCFace: Synthetic Face Generation with Dual Condition Diffusion Model},'' in \emph{Proc. IEEE/CVF Conference on Computer Vision and Pattern Recognition}, 2023.

\bibitem{boutros2023idiff}
F.~Boutros, J.~H. Grebe, A.~Kuijper, and N.~Damer, ``{IDiff-Face: Synthetic-based Face Recognition through Fizzy Identity-Conditioned Diffusion Model},'' in \emph{Proc. IEEE/CVF International Conference on Computer Vision}, 2023.

\bibitem{boutros2023synthetic}
F.~Boutros, V.~Struc, J.~Fierrez, and N.~Damer, ``{Synthetic Data for Face Recognition: Current State and Future Prospects},'' \emph{Image and Vision Computing}, vol. 135, p. 104688, 2023.

\bibitem{joshi2024synthetic}
I.~Joshi, M.~Grimmer, C.~Rathgeb, C.~Busch, F.~Bremond, and A.~Dantcheva, ``{Synthetic Data in Human Analysis: A Survey},'' \emph{IEEE Transactions on Pattern Analysis and Machine Intelligence}, vol.~46, no.~7, pp. 4957--4976, 2024.

\bibitem{sun2018demographic}
Y.~Sun, M.~Zhang, Z.~Sun, and T.~Tan, ``{Demographic Analysis from Biometric Data: Achievements, Challenges, and New Frontiers},'' \emph{IEEE Transactions on Pattern Analysis and Machine Intelligence}, vol.~40, no.~2, pp. 332--351, 2018.

\bibitem{goodfellow2014generative}
I.~Goodfellow, J.~Pouget-Abadie, M.~Mirza, B.~Xu, D.~Warde-Farley, S.~Ozair, A.~Courville, and Y.~Bengio, ``{Generative Adversarial Nets},'' in \emph{Proc. Advances in Neural Information Processing Systems}, 2014.

\bibitem{ho2020denoising}
J.~Ho, A.~Jain, and P.~Abbeel, ``{Denoising Diffusion Probabilistic Models},'' in \emph{Proc. Advances in Neural Information Processing Systems}, 2020.

\bibitem{bae2023digiface}
G.~Bae, M.~de~La~Gorce, T.~Baltru\v{s}aitis, C.~Hewitt, D.~Chen, J.~Valentin, R.~Cipolla, and J.~Shen, ``{DigiFace-1M: 1 Million Digital Face Images for Face Recognition},'' in \emph{Proc. IEEE/CVF Winter Conference on Applications of Computer Vision}, 2023.

\bibitem{zhang2023iti}
C.~Zhang, X.~Chen, S.~Chai, C.~H. Wu, D.~Lagun, T.~Beeler, and F.~De~la Torre, ``{ITI-GEN: Inclusive Text-to-Image Generation},'' in \emph{Proc. IEEE/CVF International Conference on Computer Vision}, 2023.

\bibitem{qiu2021synface}
H.~Qiu, B.~Yu, D.~Gong, Z.~Li, W.~Liu, and D.~Tao, ``{SynFace: Face Recognition with Synthetic Data},'' in \emph{Proc. IEEE/CVF International Conference on Computer Vision}, 2021.

\bibitem{melzi2024frcsyn}
P.~Melzi, R.~Tolosana, R.~Vera-Rodriguez \emph{et~al.}, ``{FRCSyn-onGoing: Benchmarking and Comprehensive Evaluation of Real and Synthetic Data to Improve Face Recognition Systems},'' \emph{Information Fusion}, vol. 107, p. 102322, 2024.

\bibitem{melzi2024frcsyna}
------, ``{FRCSyn Challenge at WACV 2024: Face Recognition Challenge in the Era of Synthetic Data},'' in \emph{Proc. of the IEEE/CVF Winter Conference on Applications of Computer Vision Workshops}, 2024.

\bibitem{he2016deep}
K.~He, X.~Zhang, S.~Ren, and J.~Sun, ``{Deep Residual Learning for Image Recognition},'' in \emph{Proc. IEEE/CVF Conference on Computer Vision and Pattern Recognition}, 2016.

\bibitem{radford2021learning}
A.~Radford, J.~W. Kim, C.~Hallacy, A.~Ramesh, G.~Goh, S.~Agarwal, G.~Sastry, A.~Askell, P.~Mishkin, J.~Clark, G.~Krueger, and I.~Sutskever, ``{Learning Transferable Visual Models From Natural Language Supervision},'' in \emph{Proc. 38th International Conference on Machine Learning}, 2021.

\bibitem{oquab2023dinov2}
M.~Oquab, T.~Darcet, T.~Moutakanni, H.~Vo, M.~Szafraniec, V.~Khalidov, P.~Fernandez, D.~Haziza, F.~Massa, A.~El-Nouby \emph{et~al.}, ``{DinoV2: Learning Robust Visual Features Without Supervision},'' \emph{arXiv preprint arXiv:2304.07193}, 2023.

\bibitem{wang2023hulk}
Y.~Wang, Y.~Wu, S.~Tang, W.~He, X.~Guo, F.~Zhu, L.~Bai, R.~Zhao, J.~Wu, T.~He \emph{et~al.}, ``{Hulk: A Universal Knowledge Translator For Human-Centric Tasks},'' \emph{arXiv preprint arXiv:2312.01697}, 2023.

\bibitem{tang2023humanbench}
S.~Tang, C.~Chen, Q.~Xie, M.~Chen, Y.~Wang, Y.~Ci, L.~Bai, F.~Zhu, H.~Yang, L.~Yi, R.~Zhao, and W.~Ouyang, ``{HumanBench: Towards General Human-Centric Perception With Projector Assisted Pretraining},'' in \emph{Proc. IEEE/CVF Conference on Computer Vision and Pattern Recognition}, 2023.

\bibitem{khirodkar2025sapiens}
R.~Khirodkar, T.~Bagautdinov, J.~Martinez, S.~Zhaoen, A.~James, P.~Selednik, S.~Anderson, and S.~Saito, ``{Sapiens: Foundation for Human Vision Models},'' in \emph{Proc. European Conference on Computer Vision}, 2025.

\bibitem{bozorgtabar2019using}
B.~Bozorgtabar, M.~S. Rad, H.~K. Ekenel, and J.-P. Thiran, ``{Using Photorealistic Face Synthesis and Domain Adaptation to Improve Facial Expression Analysis},'' in \emph{Proc. IEEE International Conference on Automatic Face \& Gesture Recognition}, 2019.

\bibitem{tolosana2021deepwritesyn}
R.~Tolosana, P.~Delgado-Santos, A.~Perez-Uribe, R.~Vera-Rodriguez, J.~Fierrez, and A.~Morales, ``{DeepWriteSYN: On-Line Handwriting Synthesis via Deep Short-Term Representations},'' \emph{Proc. AAAI Conference on Artificial Intelligence}, vol.~35, pp. 600--608, 2021.

\bibitem{hwang2023eldersim}
H.~Hwang, C.~Jang, G.~Park, J.~Cho, and I.-J. Kim, ``{ElderSim: A Synthetic Data Generation Platform for Human Action Recognition in Eldercare Applications},'' \emph{IEEE Access}, vol.~11, pp. 9279--9294, 2023.

\bibitem{varol2017learning}
G.~Varol, J.~Romero, X.~Martin, N.~Mahmood, M.~J. Black, I.~Laptev, and C.~Schmid, ``{Learning From Synthetic Humans},'' in \emph{Proc. IEEE Conference on Computer Vision and Pattern Recognition}, 2017.

\bibitem{deandres2024second}
I.~DeAndres-Tame, R.~Tolosana, P.~Melzi, R.~Vera-Rodriguez, M.~Kim, C.~Rathgeb, X.~Liu, A.~Morales, J.~Fierrez, J.~Ortega-Garcia, Z.~Zhong, Y.~Huang, Y.~Mi, S.~Ding, S.~Zhou, S.~He, L.~Fu, H.~Cong, R.~Zhang, Z.~Xiao, E.~Smirnov, A.~Pimenov, A.~Grigorev, D.~Timoshenko, K.~M. Asfaw, C.~Y. Low, H.~Liu, C.~Wang, Q.~Zuo, Z.~He, H.~O. Shahreza, A.~George, A.~Unnervik, P.~Rahimi, S.~Marcel, P.~C. Neto, M.~Huber, J.~N. Kolf, N.~Damer, F.~Boutros, J.~S. Cardoso, A.~F. Sequeira, A.~Atzori, G.~Fenu, M.~Marras, V.~Štruc, J.~Yu, Z.~Li, J.~Li, W.~Zhao, Z.~Lei, X.~Zhu, X.-Y. Zhang, B.~Biesseck, P.~Vidal, L.~Coelho, R.~Granada, and D.~Menotti, ``{Second Edition FRCSyn Challenge at CVPR 2024: Face Recognition Challenge in the Era of Synthetic Data},'' in \emph{Proc. IEEE/CVF Conference on Computer Vision and Pattern Recognition}, 2024.

\bibitem{kansy2023controllable}
M.~Kansy, A.~Ra\"el, G.~Mignone, J.~Naruniec, C.~Schroers, M.~Gross, and R.~M. Weber, ``{Controllable Inversion of Black-Box Face Recognition Models via Diffusion},'' in \emph{Proc. IEEE/CVF International Conference on Computer Vision Workshops}, 2023.

\bibitem{boutros2022sface}
F.~Boutros, M.~Huber, P.~Siebke, T.~Rieber, and N.~Damer, ``{SFace: Privacy-friendly and Accurate Face Recognition using Synthetic Data},'' in \emph{Proc. IEEE International Joint Conference on Biometrics}, 2022.

\bibitem{karras2021alias}
T.~Karras, M.~Aittala, S.~Laine, E.~H{\"a}rk{\"o}nen, J.~Hellsten, J.~Lehtinen, and T.~Aila, ``{Alias-Free Generative Adversarial Networks},'' \emph{Advances in Neural Information Processing Systems}, vol.~34, pp. 852--863, 2021.

\bibitem{ruiz2023dreambooth}
N.~Ruiz, Y.~Li, V.~Jampani, Y.~Pritch, M.~Rubinstein, and K.~Aberman, ``{DreamBooth: Fine Tuning Text-to-Image Diffusion Models for Subject-Driven Generation},'' in \emph{Proc. IEEE/CVF Conference on Computer Vision and Pattern Recognition}, 2023.

\bibitem{wood2021fake}
E.~Wood, T.~Baltru\v{s}aitis, C.~Hewitt, S.~Dziadzio, T.~J. Cashman, and J.~Shotton, ``{Fake It Till You Make It: Face Analysis in the Wild Using Synthetic Data Alone},'' in \emph{Proc. IEEE/CVF International Conference on Computer Vision}, 2021.

\bibitem{deng2020disentangled}
Y.~Deng, J.~Yang, D.~Chen, F.~Wen, and X.~Tong, ``{Disentangled and Controllable Face Image Generation via 3D Imitative-Contrastive Learning},'' in \emph{Proc. IEEE/CVF Conference on Computer Vision and Pattern Recognition}, 2020.

\bibitem{yi2014learning}
D.~Yi, Z.~Lei, S.~Liao, and S.~Z. Li, ``{Learning Face Representation from Scratch},'' \emph{arXiv preprint arXiv:1411.7923}, 2014.

\bibitem{wang2020mitigating}
M.~Wang and W.~Deng, ``{Mitigating Bias in Face Recognition Using Skewness-Aware Reinforcement Learning},'' in \emph{Proc. IEEE/CVF Conference on Computer Vision and Pattern Recognition}, 2020.

\bibitem{moschoglou2017agedb}
S.~Moschoglou, A.~Papaioannou, C.~Sagonas, J.~Deng, I.~Kotsia, and S.~Zafeiriou, ``{AgeDB: The First Manually Collected, In-The-Wild Age Database},'' in \emph{Proc. IEEE/CVF Conference on Computer Vision and Pattern Recognition Workshops}, 2017.

\bibitem{sengupta2016frontal}
S.~Sengupta, J.-C. Chen, C.~Castillo, V.~M. Patel, R.~Chellappa, and D.~W. Jacobs, ``{Frontal to Profile Face Verification in the Wild},'' in \emph{Proc. IEEE/CVF Winter Conference on Applications of Computer Vision}, 2016.

\bibitem{erakιn2021recognizing}
M.~E. Erak$\iota$n, U.~Demir, and H.~K. Ekenel, ``{On Recognizing Occluded Faces in the Wild},'' in \emph{Proc. International Conference of the Biometrics Special Interest Group}, 2021.

\bibitem{karkkainen2021fairface}
K.~Karkkainen and J.~Joo, ``{FairFace: Face Attribute Dataset for Balanced Race, Gender, and Age for Bias Measurement and Mitigation},'' in \emph{Proc. IEEE/CVF Winter Conference on Applications of Computer Vision}, 2021.

\bibitem{duta2021improved}
I.~C. Duta, L.~Liu, F.~Zhu, and L.~Shao, ``{Improved Residual Networks for Image and Video Recognition},'' in \emph{Proc. 25th International Conference on Pattern Recognition}, 2021.

\bibitem{rombach2022high}
R.~Rombach, A.~Blattmann, D.~Lorenz, P.~Esser, and B.~Ommer, ``{High-Resolution Image Synthesis With Latent Diffusion Models},'' in \emph{Proceedings of the IEEE/CVF Conference on Computer Vision and Pattern Recognition}, 2022.

\bibitem{song2021denoising}
J.~Song, C.~Meng, and S.~Ermon, ``{Denoising Diffusion Implicit Models},'' in \emph{Proc. International Conference on Learning Representations}, 2021.

\bibitem{li2024photomaker}
Z.~Li, M.~Cao, X.~Wang, Z.~Qi, M.-M. Cheng, and Y.~Shan, ``{PhotoMaker: Customizing Realistic Human Photos via Stacked ID Embedding},'' in \emph{Proc. IEEE/CVF Conference on Computer Vision and Pattern Recognition}, 2024.

\bibitem{crowson2024scalable}
K.~Crowson, S.~A. Baumann, A.~Birch, T.~M. Abraham, D.~Z. Kaplan, and E.~Shippole, ``{Scalable High-Resolution Pixel-Space Image Synthesis with Hourglass Diffusion Transformers},'' \emph{arXiv preprint arXiv:2401.11605}, 2024.

\bibitem{oord2017neural}
A.~van~den Oord, O.~Vinyals, and k.~kavukcuoglu, ``{Neural Discrete Representation Learning},'' in \emph{Proc. Advances in Neural Information Processing Systems}, 2017.

\bibitem{walton2022stylenat}
S.~Walton, A.~Hassani, X.~Xu, Z.~Wang, and H.~Shi, ``{StyleNAT: Giving Each Head a New Perspective},'' \emph{arXiv preprint arXiv:2211.05770}, 2022.

\bibitem{garaev2023facemixa}
N.~Garaev, E.~Smirnov, V.~Galyuk, and E.~Lukyanets, ``{FaceMix: Transferring Local Regions for Data Augmentation in Face Recognition},'' in \emph{Proc. Neural Information Processing}, 2023.

\bibitem{zhou2023uniface}
J.~Zhou, X.~Jia, Q.~Li, L.~Shen, and J.~Duan, ``{UniFace: Unified Cross-Entropy Loss for Deep Face Recognition},'' in \emph{Proc. IEEE/CVF International Conference on Computer Vision}, 2023.

\bibitem{low2023slackedface}
C.~Y. Low, J.~C.~L. Chai, J.~Park, K.~Ann, and M.~Cha, ``{SlackedFace: Learning a Slacked Margin for Low-Resolution Face Recognition},'' in \emph{Proc. 34th British Machine Vision Conference}, 2023.

\bibitem{hu2018squeeze}
J.~Hu, L.~Shen, and G.~Sun, ``{Squeeze-and-Excitation Networks},'' in \emph{Proc. IEEE Conference on Computer Vision and Pattern Recognition}, 2018.

\bibitem{wang2018cosface}
H.~Wang, Y.~Wang, Z.~Zhou, X.~Ji, D.~Gong, J.~Zhou, Z.~Li, and W.~Liu, ``{CosFace: Large Margin Cosine Loss for Deep Face Recognition},'' in \emph{Proc. IEEE Conference on Computer Vision and Pattern Recognition}, 2018.

\bibitem{karras2019style}
T.~Karras, S.~Laine, and T.~Aila, ``{A Style-Based Generator Architecture for Generative Adversarial Networks},'' in \emph{Proc. IEEE/CVF conference on Computer Vision and Pattern Recognition}, 2019.

\bibitem{otroshishahreza2024knowledge}
H.~Otroshi~Shahreza, A.~George, and S.~Marcel, ``{Knowledge Distillation for Face Recognition Using Synthetic Data With Dynamic Latent Sampling},'' \emph{IEEE Access}, vol.~12, pp. 187\,800--187\,812, 2024.

\bibitem{boutros2022elasticface}
F.~Boutros, N.~Damer, F.~Kirchbuchner, and A.~Kuijper, ``{ElasticFace: Elastic Margin Loss for Deep Face Recognition},'' in \emph{Proc. IEEE/CVF Conference on Computer Vision and Pattern Recognition Workshops}, 2022.

\bibitem{boutros2023exfacegan}
F.~Boutros, M.~Klemt, M.~Fang, A.~Kuijper, and N.~Damer, ``{ExFaceGAN: Exploring Identity Directions in GAN’s Learned Latent Space for Synthetic Identity Generation},'' in \emph{IEEE International Joint Conference on Biometrics}, 2023.

\bibitem{an2021partial}
X.~An, X.~Zhu, Y.~Gao, Y.~Xiao, Y.~Zhao, Z.~Feng, L.~Wu, B.~Qin, M.~Zhang, D.~Zhang, and Y.~Fu, ``{Partial FC: Training 10 Million Identities on a Single Machine},'' in \emph{Proc. IEEE/CVF International Conference on Computer Vision Workshops}, 2021.

\bibitem{zhu2021webface260m}
Z.~Zhu, G.~Huang, J.~Deng, Y.~Ye, J.~Huang, X.~Chen, J.~Zhu, T.~Yang, J.~Lu, D.~Du, and J.~Zhou, ``{WebFace260M: A Benchmark Unveiling the Power of Million-Scale Deep Face Recognition},'' in \emph{Proc. IEEE/CVF Conference on Computer Vision and Pattern Recognition}, 2021.

\bibitem{ronneberger2015u}
O.~Ronneberger, P.~Fischer, and T.~Brox, ``{U-Net: Convolutional Networks for Biomedical Image Segmentation},'' in \emph{Proc. Medical image computing and computer-assisted intervention}, 2015.

\bibitem{tong2024improving}
A.~Tong, K.~FATRAS, N.~Malkin, G.~Huguet, Y.~Zhang, J.~Rector-Brooks, G.~Wolf, and Y.~Bengio, ``{Improving and Generalizing Flow-Based Generative Models with Minibatch Optimal Transport},'' \emph{Transactions on Machine Learning Research}, 2024.

\bibitem{ho2022classifier}
J.~Ho and T.~Salimans, ``{Classifier-Free Diffusion Guidance},'' \emph{arXiv preprint arXiv:2207.12598}, 2022.

\bibitem{smirnov2022prototype}
E.~Smirnov, N.~Garaev, V.~Galyuk, and E.~Lukyanets, ``{Prototype Memory for Large-Scale Face Representation Learning},'' \emph{IEEE Access}, vol.~10, pp. 12\,031--12\,046, 2022.

\bibitem{deng2020retinaface}
J.~Deng, J.~Guo, E.~Ververas, I.~Kotsia, and S.~Zafeiriou, ``{RetinaFace: Single-Shot Multi-Level Face Localisation in the Wild},'' in \emph{Proc. IEEE/CVF Conference on Computer Vision and Pattern Recognition}, 2020.

\bibitem{karras2020analyzing}
T.~Karras, S.~Laine, M.~Aittala, J.~Hellsten, J.~Lehtinen, and T.~Aila, ``{Analyzing and Improving the Image Quality of StyleGAN},'' in \emph{Proc. IEEE/CVF Conference on Computer Vision and Pattern Recognition}, 2020.

\bibitem{neto2023compressed}
P.~C. Neto, E.~Caldeira, J.~S. Cardoso, and A.~F. Sequeira, ``{Compressed Models Decompress Race Biases: What Quantized Models Forget for Fair Face Recognition},'' in \emph{Proc. International Conference of the Biometrics Special Interest Group}, 2023.

\bibitem{neto2022ocfr}
P.~C. Neto, F.~Boutros, J.~R. Pinto, N.~Damer, A.~F. Sequeira, J.~S. Cardoso, M.~Bengherabi, A.~Bousnat, S.~Boucheta, N.~Hebbadj, M.~E. Erakın, U.~Demir, H.~K. Ekenel, P.~B. De~Queiroz~Vidal, and D.~Menotti, ``{OCFR 2022: Competition on Occluded Face Recognition from Synthetically Generated Structure-Aware Occlusions},'' in \emph{Proc. IEEE International Joint Conference on Biometrics}, 2022.

\bibitem{neto2024massively}
P.~C. Neto, R.~M. Mamede, C.~Albuquerque, T.~Gon{\c{c}}alves, and A.~F. Sequeira, ``{Massively Annotated Datasets for Assessment of Synthetic and Real Data in Face Recognition},'' \emph{arXiv preprint arXiv:2404.15234}, 2024.

\bibitem{atzori2024if}
A.~Atzori, F.~Boutros, N.~Damer, G.~Fenu, and M.~Marras, ``{If It's Not Enough, Make It So: Reducing Authentic Data Demand in Face Recognition through Synthetic Faces},'' in \emph{Proc. 18th International Conference on Automatic Face and Gesture Recognition}, 2024.

\bibitem{shoshan2021gan}
A.~Shoshan, N.~Bhonker, I.~Kviatkovsky, and G.~Medioni, ``{GAN-Control: Explicitly Controllable GANs},'' in \emph{Proc. IEEE/CVF International Conference on Computer Vision}, 2021.

\bibitem{boutros2023unsupervised}
F.~Boutros, M.~Klemt, M.~Fang, A.~Kuijper, and N.~Damer, ``{Unsupervised Face Recognition using Unlabeled Synthetic Data},'' in \emph{Proc. 17th International Conference on Automatic Face and Gesture Recognition}, 2023.

\bibitem{zhang2016joint}
K.~Zhang, Z.~Zhang, Z.~Li, and Y.~Qiao, ``{Joint Face Detection and Alignment Using Multitask Cascaded Convolutional Networks},'' \emph{IEEE Signal Processing Letters}, vol.~23, no.~10, pp. 1499--1503, 2016.

\bibitem{wang2021facex}
J.~Wang, Y.~Liu, Y.~Hu, H.~Shi, and T.~Mei, ``{FaceX-Zoo: A PyTorch Toolbox for Face Recognition},'' in \emph{Proc. 29th ACM International Conference on Multimedia}, 2021.

\bibitem{ngan2020ongoing}
M.~L. Ngan, P.~J. Grother, and K.~K. Hanaoka, ``{OnGoing Face Recognition Vendor Test (FRVT) Part 6b: Face Recognition Accuracy with Face Masks Using Post-Covid-19 Algorithms},'' 2020.

\bibitem{dealcala2024is}
D.~DeAlcala, A.~Morales, G.~Mancera, J.~Fierrez, R.~Tolosana, and J.~Ortega-Garcia, ``{Is my Data in your AI Model? Membership Inference Test with Application to Face Images},'' \emph{arXiv preprint arXiv:2402.09225}, 2024.

\bibitem{shahreza2024sdfr}
H.~O. Shahreza, C.~Ecabert, A.~George, A.~Unnervik, S.~Marcel, N.~D. Domenico, G.~Borghi, D.~Maltoni, F.~Boutros, J.~Vogel, N.~Damer, Ángela Sánchez-Pérez, EnriqueMas-Candela, J.~Calvo-Zaragoza, B.~Biesseck, P.~Vidal, R.~Granada, D.~Menotti, I.~DeAndres-Tame, S.~M.~L. Cava, S.~Concas, P.~Melzi, R.~Tolosana, R.~Vera-Rodriguez, G.~Perelli, G.~Orrù, G.~L. Marcialis, and J.~Fierrez, ``{SDFR: Synthetic Data for Face Recognition Competition},'' in \emph{Proc. 18th IEEE International Conference on Automatic Face and Gesture Recognition}, 2024.

\bibitem{liu2024farsight}
F.~Liu, R.~Ashbaugh, N.~Chimitt, N.~Hassan, A.~Hassani, A.~Jaiswal, M.~Kim, Z.~Mao, C.~Perry, Z.~Ren, Y.~Su, P.~Varghaei, K.~Wang, X.~Zhang, S.~Chan, A.~Ross, H.~Shi, Z.~Wang, A.~Jain, and X.~Liu, ``{FarSight: A Physics-Driven Whole-Body Biometric System at Large Distance and Altitude},'' in \emph{Proc. IEEE/CVF Winter Conference on Applications of Computer Vision}, 2024.

\end{thebibliography}

\section{Biography Section}
\vskip -2\baselineskip plus -1fil
\begin{IEEEbiographynophoto}{Ivan DeAndres-Tame} is a Ph.D. at Universidad Autónoma de Madrid at Madrid, Spain. Contact him at ivan.deandres@uam.es.\end{IEEEbiographynophoto}
\vskip -2\baselineskip plus -1fil
\begin{IEEEbiographynophoto}{Ruben Tolosana} is an associate professor at Universidad Autónoma de Madrid at Madrid, Spain. Contact him at ruben.tolosana@uam.es.\end{IEEEbiographynophoto}
\vskip -2\baselineskip plus -1fil
\begin{IEEEbiographynophoto}{Pietro Melzi} got his Ph.D. at Universidad Autónoma de Madrid at Madrid, Spain. Contact him at pietro.melzi@uam.es. \end{IEEEbiographynophoto}
\vskip -2\baselineskip plus -1fil
\begin{IEEEbiographynophoto}{Ruben Vera-Rodriguez} is an associate professor at Universidad Autónoma de Madrid at Madrid, Spain. Contact him at ruben.vera@uam.es.\end{IEEEbiographynophoto}
\vskip -2\baselineskip plus -1fil
\begin{IEEEbiographynophoto}{Minchul Kim} is a Ph.D. at Michigan State University, at Michigan, USA. Contact him at kimminc2@msu.edu.\end{IEEEbiographynophoto}
\vskip -2\baselineskip plus -1fil
\begin{IEEEbiographynophoto}{Christian Rathgeb} is a full professor at  Hochschule Darmstadt at Darmstadt, Germany. Contact him at christian.rathgeb@h-da.de.\end{IEEEbiographynophoto}
\vskip -2\baselineskip plus -1fil
\begin{IEEEbiographynophoto}{Xiaoming Liu} is a full professor at Michigan State University, at Michigan, USA. Contact him at liuxm@cse.msu.edu.\end{IEEEbiographynophoto}
\vskip -2\baselineskip plus -1fil
\begin{IEEEbiographynophoto}{Luis F.Gomez} got his Ph.D. at Universidad Autónoma de Madrid at Madrid, Spain. Contact him at luisf.gomez@uam.es. \end{IEEEbiographynophoto}
\vskip -2\baselineskip plus -1fil
\begin{IEEEbiographynophoto}{Aythami Morales} is an associate professor at Universidad Autónoma de Madrid at Madrid, Spain. Contact him at aythami.morales@uam.es.\end{IEEEbiographynophoto}
\vskip -2\baselineskip plus -1fil
\begin{IEEEbiographynophoto}{Julian Fierrez} is a full professor at Universidad Autónoma de Madrid at Madrid, Spain. Contact him at rjulian.fierrez@uam.es.\end{IEEEbiographynophoto}
\vskip -2\baselineskip plus -1fil
\begin{IEEEbiographynophoto}{Javier Ortega-Garcia} is a full professor at Universidad Autónoma de Madrid at Madrid, Spain. Contact him at javier.ortega@uam.es.\end{IEEEbiographynophoto}
\vskip -2\baselineskip plus -1fil
\begin{IEEEbiographynophoto}{Zhizhou Zhong} is an M.S. student at Fudan University at Shanghai, China. Contact him at zzzhong22@m.fudan.edu.cn.\end{IEEEbiographynophoto}
\vskip -2\baselineskip plus -1fil
\begin{IEEEbiographynophoto}{Yuge Huang} is a senior researcher at Tencent Youtu Lab at Shanghai, China. Contact him at yugehuang@tencent.com.\end{IEEEbiographynophoto}
\vskip -2\baselineskip plus -1fil
\begin{IEEEbiographynophoto}{Yuxi Mi} is a Ph.D. at Fudan University at Shanghai, China. Contact him at yxmi20@fudan.edu.cn.\end{IEEEbiographynophoto}
\vskip -2\baselineskip plus -1fil
\begin{IEEEbiographynophoto}{Shouhong Ding} is a principal researcher at Tencent Youtu Lab at Shanghai, China. Contact him at ericshding@tencent.com.\end{IEEEbiographynophoto}
\vskip -2\baselineskip plus -1fil
\begin{IEEEbiographynophoto}{Shuigeng Zhou} is a full professor at Fudan University at Shanghai, China. Contact him at sgzhou@fudan.edu.cn.\end{IEEEbiographynophoto}
\vskip -2\baselineskip plus -1fil
\begin{IEEEbiographynophoto}{Shuai He} is a Software Engineer at the Interactive Entertainment Group of Netease Inc at Guangzhou, China. Contact him at heshuai03@corp.netease.com.\end{IEEEbiographynophoto}
\vskip -2\baselineskip plus -1fil
\begin{IEEEbiographynophoto}{Lingzhi Fu} is a Software Engineer at the Interactive Entertainment Group of Netease Inc at Guangzhou, China. Contact him at fulingzhi@corp.netease.com.\end{IEEEbiographynophoto}
\vskip -2\baselineskip plus -1fil
\begin{IEEEbiographynophoto}{Heng Cong} is a Software Engineer at the Interactive Entertainment Group of Netease Inc at Guangzhou, China. Contact him at congheng@corp.netease.com.\end{IEEEbiographynophoto}
\vskip -2\baselineskip plus -1fil
\begin{IEEEbiographynophoto}{Rongyu Zhang} is a Software Engineer at the Interactive Entertainment Group of Netease Inc at Guangzhou, China. Contact her at zhangrongyu@corp.netease.com.\end{IEEEbiographynophoto}
\vskip -2\baselineskip plus -1fil
\begin{IEEEbiographynophoto}{Zhihong Xiao} is a Software Engineer at the Interactive Entertainment Group of Netease Inc at Guangzhou, China. Contact him at xiaozhihong@corp.netease.com.\end{IEEEbiographynophoto}
\vskip -2\baselineskip plus -1fil
\begin{IEEEbiographynophoto}{Evgeny Smirnov} is a ML engineer at ID R\&D Inc. in Barcelona, Spain. Contact him at evgeny.smirnov@idrnd.net.\end{IEEEbiographynophoto}
\vskip -2\baselineskip plus -1fil
\begin{IEEEbiographynophoto}{Anton Pimenov} is a ML engineer at ID R\&D Inc. in Barcelona, Spain. Contact him at pimenov@idrnd.net.\end{IEEEbiographynophoto}
\vskip -2\baselineskip plus -1fil
\begin{IEEEbiographynophoto}{Aleksei Grigorev} is a ML Engineer at ID R\&D Inc. in Barcelona, Spain. Contact him at grigoriev@idrnd.net.\end{IEEEbiographynophoto}
\vskip -2\baselineskip plus -1fil
\begin{IEEEbiographynophoto}{Denis Timoshenko} is a ML engineer at ID R\&D Inc. in New York, USA. Contact him at timoshenko@idrnd.net.\end{IEEEbiographynophoto}
\vskip -2\baselineskip plus -1fil
\begin{IEEEbiographynophoto}{Kaleb Mesfin Asfaw} is an undergraduate student researcher with the Korea Advanced Institute of Science and Technology at Daejeon, South Korea. Contact him at kalebmes06@kaist.ac.kr.\end{IEEEbiographynophoto}
\vskip -2\baselineskip plus -1fil
\begin{IEEEbiographynophoto}{Cheng Yaw Low} is a research associate with the Institute for Basic Science at Daejeon, South Korea. Contact him at chengyawlow@ibs.re.kr.\end{IEEEbiographynophoto}
\vskip -2\baselineskip plus -1fil
\begin{IEEEbiographynophoto}{Hao Liu} is a M.S at China Telecom AI, China. Contact him at liuh9@chinatelecom.cn\end{IEEEbiographynophoto}
\vskip -2\baselineskip plus -1fil
\begin{IEEEbiographynophoto}{Chuyi Wang} is a M.S at China Telecom AI, China. Contact him at wangcy30@chinatelecom.cn.\end{IEEEbiographynophoto}
\vskip -2\baselineskip plus -1fil
\begin{IEEEbiographynophoto}{Qing Zuo} is a Ph.D. at China Telecom AI, China. Contact him at zuoq2@chinatelecom.cn\end{IEEEbiographynophoto}
\vskip -2\baselineskip plus -1fil
\begin{IEEEbiographynophoto}{Zhixiang He} is a Ph.D. at China Telecom AI, China. Contact him at hezx3@chinatelecom.cn.\end{IEEEbiographynophoto}
\vskip -2\baselineskip plus -1fil
\begin{IEEEbiographynophoto}{Hatef Otroshi Shahreza} is a Ph.D. student at EPFL at Lausanne, Switzerland and a Research Assistant at Idiap Research Institute at Martigny, Switzerland. Contact him at hatef.otroshi@idiap.ch.\end{IEEEbiographynophoto}
\vskip -2\baselineskip plus -1fil
\begin{IEEEbiographynophoto}{Anjith George} is a Research Associate at Idiap Research Institute at Martigny, Switzerland. Contact him at anjith.george@idiap.ch.\end{IEEEbiographynophoto}
\vskip -2\baselineskip plus -1fil
\begin{IEEEbiographynophoto}{Alexander Unnervik} is a Ph.D. student at EPFL at Lausanne, Switzerland and a Research Assistant at Idiap Research Institute at Martigny, Switzerland.  Contact him at alex.unnervik@idiap.ch.\end{IEEEbiographynophoto}
\vskip -2\baselineskip plus -1fil
\begin{IEEEbiographynophoto}{Parsa Rahimi} is a Ph.D. student at EPFL at Lausanne, Switzerland and a Research Assistant at Idiap Research Institute at Martigny, Switzerland.  Contact him at parsa.rahimi@idiap.ch.\end{IEEEbiographynophoto}
\vskip -2\baselineskip plus -1fil
\begin{IEEEbiographynophoto}{Sébastien Marcel} is a Senior Researcher and the Head of Biometrics Security and Privacy group at Idiap Research Institute at Martigny, Switzerland and a Professor at UNIL at Lausanne, Switzerland.  Contact him at sebastien.marcel@idiap.ch.\end{IEEEbiographynophoto}
\vskip -2\baselineskip plus -1fil
\begin{IEEEbiographynophoto}{Pedro C. Neto} is a research assistant at INESC TEC at Porto, Portugal. Contact him at Pedro.d.carneiro@inesctec.pt\end{IEEEbiographynophoto}
\vskip -2\baselineskip plus -1fil
\begin{IEEEbiographynophoto}{Marco Huber} is a research associate at Fraunhofer IGD at Darmstadt, Germany. Contact him at marco.huber@igd.fraunhofer.de\end{IEEEbiographynophoto}
\vskip -2\baselineskip plus -1fil
\begin{IEEEbiographynophoto}{Jan Niklas Kolf} is a research associate at Fraunhofer IGD at Darmstadt, Germany. Contact him at jan.niklas.kolf@igd.fraunhofer.de\end{IEEEbiographynophoto}
\vskip -2\baselineskip plus -1fil
\begin{IEEEbiographynophoto}{Naser Damer} is a research associate at Fraunhofer IGD at Darmstadt, Germany. Contact him at naser.damer@igd.fraunhofer.de\end{IEEEbiographynophoto}
\vskip -2\baselineskip plus -1fil
\begin{IEEEbiographynophoto}{Fadi Boutros} is a research associate at Fraunhofer IGD at Darmstadt, Germany. Contact him at fadi.boutros@igd.fraunhofer.de.\end{IEEEbiographynophoto}
\vskip -2\baselineskip plus -1fil
\begin{IEEEbiographynophoto}{Jaime S. Cardoso} is a full professor at FEUP at Porto, Portugal. Contact him at jaime.cardoso@fe.up.pt\end{IEEEbiographynophoto}
\vskip -2\baselineskip plus -1fil
\begin{IEEEbiographynophoto}{Ana F. Sequeira} is an assistant researcher at INESC TEC at Porto, Portugal. Contact her at ana.f.sequeira@inesctec.pt\end{IEEEbiographynophoto}
\vskip -2\baselineskip plus -1fil
\begin{IEEEbiographynophoto}{Andrea Atzori} is a Ph.D. student at University of Cagliari at Cagliari, Italy. Contact him at andrea.atzori@unica.it.\end{IEEEbiographynophoto}
\vskip -2\baselineskip plus -1fil
\begin{IEEEbiographynophoto}{Gianni Fenu} is a Ph.D. student at University of Cagliari at Cagliari, Italy. Contact him at gianni.fenu@unica.it.\end{IEEEbiographynophoto}
\vskip -2\baselineskip plus -1fil
\begin{IEEEbiographynophoto}{Mirko Marras} is a Ph.D. student at University of Cagliari at Cagliari, Italy. Contact him at mirko.marras@acm.org.\end{IEEEbiographynophoto}
\vskip -2\baselineskip plus -1fil
\begin{IEEEbiographynophoto}{Vitomir \v{S}truc} is a Full Professors at the University of Ljubljana at Ljubljana, Slovenia. Contact him at vitomir.struc@fe.uni-lj.si\end{IEEEbiographynophoto}
\vskip -2\baselineskip plus -1fil
\begin{IEEEbiographynophoto}{Jiang Yu} is a engineer at Samsung Electronics (China) R\&D Centre, China. Contact him at jiang0922.yu@samsung.com.\end{IEEEbiographynophoto}
\vskip -2\baselineskip plus -1fil
\begin{IEEEbiographynophoto}{Zhangjie Li} is a student at University of Science and Technology of China, China. Contact him at lizhangjie@mail.ustc.edu.\end{IEEEbiographynophoto}
\vskip -2\baselineskip plus -1fil
\begin{IEEEbiographynophoto}{Jichun Li} is a engineer at Samsung Electronics (China) R\&D Centre, China. Contact him at jichun.li@samsung.com.\end{IEEEbiographynophoto}
\vskip -2\baselineskip plus -1fil
\begin{IEEEbiographynophoto}{Weisong Zhao} is a student at Institute of Information Engineering, Chinese Academy of Sciences at Beijing, China. Contact him at zhaoweisong@iie.ac.cn.\end{IEEEbiographynophoto}
\vskip -2\baselineskip plus -1fil
\begin{IEEEbiographynophoto}{Zhen Lei} is a professor at MAIS, Institute of Automation, Chinese Academy of Sciences at Beijing, China. Contact him at zlei@nlpr.ia.ac.cn.\end{IEEEbiographynophoto}
\vskip -2\baselineskip plus -1fil
\begin{IEEEbiographynophoto}{Xiangyu Zhu} is an associated professor at Institute of Automation, Chinese Academy of Sciences at Beijing, China. Contact him at xiangyu.zhu@ia.ac.cn\end{IEEEbiographynophoto}
\vskip -2\baselineskip plus -1fil
\begin{IEEEbiographynophoto}{Xiao-Yu Zhang}is a professor at Institute of Information Engineering, Chinese Academy of Sciences at Beijing, China. Contact him at zhangxiaoyu@iie.ac.cn \end{IEEEbiographynophoto}
\vskip -2\baselineskip plus -1fil
\begin{IEEEbiographynophoto}{Bernardo Biesseck} is a Ph.D. student at Federal University of Paraná, at Curitiba, Brazil. Contact him at bjgbiesseck@inf.ufpr.br.\end{IEEEbiographynophoto}
\vskip -2\baselineskip plus -1fil
\begin{IEEEbiographynophoto}{Pedro Vidal} is an Undergraduate student at Federal University of Parana, at Curitiba, Brazil. Contact him at pbqv20@inf.ufpr.br.\end{IEEEbiographynophoto}
\vskip -2\baselineskip plus -1fil
\begin{IEEEbiographynophoto}{Luiz Coelho} is an ML Engineer at Unico IDTech at Belo Horizonte, Brazil. Contact him at luiz.coelho@unicio.io.\end{IEEEbiographynophoto}
\vskip -2\baselineskip plus -1fil
\begin{IEEEbiographynophoto}{Roger Granada} is a Ph.D./ML Engineer at Unico IDTech at Porto Alegre, Brazil. Contact him at roger.granada@unico.io.\end{IEEEbiographynophoto}
\vskip -2\baselineskip plus -1fil
\begin{IEEEbiographynophoto}{David Menotti} is an Associate Professor at Federal University of Paraná, at Curitiba, Brazil. Contact him at menotti@inf.ufpr.br.\end{IEEEbiographynophoto}
\vfill

\end{document}